%% file: main.tex
\documentclass{article}

\usepackage[preprint]{corl_2026} 

\usepackage{booktabs}
\usepackage{makecell}
\usepackage{multirow}

\usepackage{graphicx}
\usepackage{caption}
\usepackage{placeins}
\usepackage{afterpage}

\usepackage{amsmath}
\usepackage{amssymb}
\usepackage{tabularx}
\usepackage{array}
\newcolumntype{P}[1]{>{\raggedright\arraybackslash}p{#1}}

\hypersetup{
    pdftitle={Robot-DIFT: Correspondence-Sensitive Diffusion Features for Contact-Rich Robot Manipulation},
    pdfauthor={Yu Deng, Yufeng Jin, Xiaogang Jia, Jiahong Xue, Gerhard Neumann, Georgia Chalvatzaki},
    pdfkeywords={Visuomotor Control; Diffusion Models; Manipulation; Robot Learning}
}

\title{Robot-DIFT: Correspondence-Sensitive Diffusion Features for Contact-Rich Robot Manipulation}

\author{
Yu Deng$^{1*}$ \And
Yufeng Jin$^{1,6*}$ \And
Xiaogang Jia$^{2}$ \And
Jiahong Xue$^{1}$ \AND
Gerhard Neumann$^{2,3}$ \And
Georgia Chalvatzaki$^{1,4,5}$ \AND
\normalfont $^{1}$TU Darmstadt \quad
$^{2}$KIT \quad
$^{3}$FZI \quad
$^{4}$Hessian.AI \\
$^{5}$Robotics Institute Germany \quad
$^{6}$Honda Research Institute Europe GmbH \\
$^{*}$Equal contribution
}

\begin{document}
\maketitle

\begin{abstract}
Robot manipulation often fails in the final millimeters: a policy may recognize the right object yet miss the pose offsets, boundaries, or pre-contact alignments needed for action.
We argue that such failures arise when semantic invariance suppresses correspondence cues for closed-loop control, or when these cues are not exposed to the policy in a usable form.
Modern visual encoders provide strong semantic abstractions, but contact-rich manipulation requires \textit{correspondence sensitivity}: discriminative feature responses to action-relevant changes in pose, boundary, and contact geometry.
Diffusion features provide a strong prior for dense correspondence, but direct use is impractical due to stochasticity, latency, and representation drift.
We introduce \textit{Robot-DIFT}, a deterministic diffusion-derived backbone for real-time control.
Through \emph{Manifold Distillation}, Robot-DIFT converts a noise-conditioned diffusion Teacher into a clean-input, single-pass Student while preserving the teacher's feature manifold.
A Spatial--Semantic Feature Pyramid Network (S2-FPN) fuses coarse-to-fine Student decoder features into visual tokens that expose semantic context and fine contact detail to the policy.
Across RoboCasa, LIBERO-10, and real robots, Robot-DIFT outperforms vision--language, self-supervised, geometry-oriented, and diffusion baselines on contact-sensitive tasks.
Controlled backbone/readout swaps show that S2-FPN unlocks, rather than replaces, the diffusion correspondence prior.
\end{abstract}

\afterpage{%
\begin{figure}[t!]
    \centering
    \includegraphics[width=\textwidth]{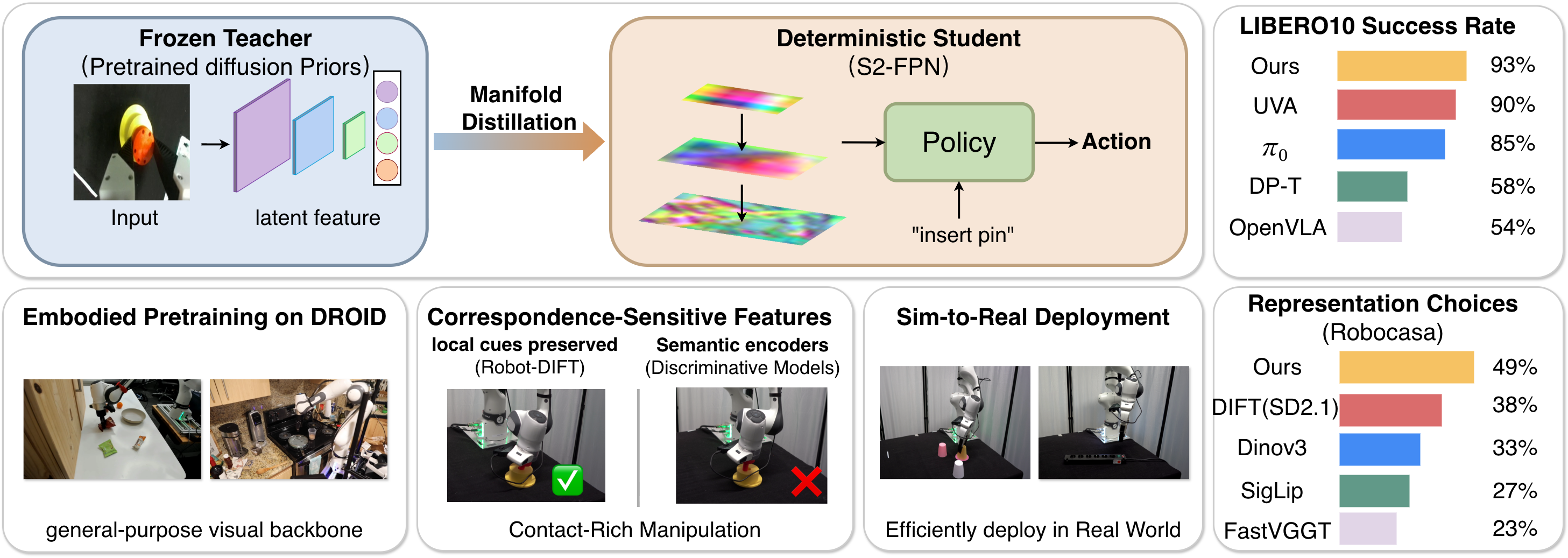}
    \caption{\textbf{Robot-DIFT distills diffusion correspondence into a deployable, contact-sensitive robot backbone.}
    Semantic encoders stay invariant to the small, contact-relevant cues that precision manipulation needs, while diffusion features preserve them but are non-deterministic and slow to deploy.
    Robot-DIFT distills a frozen diffusion Teacher into a clean-input deterministic Student pretrained on DROID, exposes its multi-scale features via S2-FPN, and drops the Teacher after training---yielding a single-pass backbone that preserves local pre-contact cues and attains the best success on LIBERO-10 and RoboCasa.}
\label{fig:robot-dift}
\vspace{-1.35em}
\end{figure}
}

\input{sections/1_introduction}
\input{sections/3_method}
\input{sections/4_experiment}
\input{sections/5_conclusion}

\bibliography{references}

\clearpage
\section*{Appendix}
\renewcommand{\thefigure}{S\arabic{figure}}
\renewcommand{\thetable}{S\arabic{table}}
\renewcommand{\theequation}{S\arabic{equation}}
\renewcommand{\thesection}{S\arabic{section}}
\renewcommand{\theHfigure}{appendix.figure.\arabic{figure}}
\renewcommand{\theHtable}{appendix.table.\arabic{table}}
\renewcommand{\theHequation}{appendix.equation.\arabic{equation}}
\renewcommand{\theHsection}{appendix.section.\arabic{section}}
\setcounter{equation}{0}
\setcounter{figure}{0}
\setcounter{table}{0}
\setcounter{section}{0}
\input{sections/appendix}

\end{document}

%% file: sections/1_introduction.tex
\section{Introduction}
\label{sec:intro}

Contact-rich manipulation often fails in the final millimeters: a policy may reach the right object, yet miss a peg-in-hole, slip during insertion, or repeatedly ``tap'' a surface without forming stable contact~\cite{xue2025reactive, hoeg2024streaming}.
These failures suggest the policy is not merely missing object identity, but acting on visual features weakly sensitive to small, contact-relevant changes in pose, boundary, and pre-contact alignment.
It perceives \emph{what} to interact with, yet lacks a representation that varies \emph{predictably} with \emph{where} and \emph{how} contact should be established.

A key bottleneck is therefore the visual representation exposed to closed-loop control.
Backbones inherit different biases from their pretraining objectives: vision--language models such as CLIP~\cite{radford2021learning} and SigLIP~\cite{zhai2023sigmoid} emphasize semantic alignment, while self-supervised recognition models such as DINOv2~\cite{oquab2023dinov2} emphasize robust object- and scene-level discrimination.
These objectives provide strong semantic priors, but can also encourage invariance to small object offsets, local boundaries, and pre-contact alignments that precision manipulation must react to~\cite{majumdar2023we, karamcheti2023language}.

We refer to this missing property as \emph{correspondence sensitivity}: behaviorally distinct but nearby object configurations should induce spatially local and discriminative feature changes.
Unlike metric 3D reconstruction, correspondence sensitivity requires policy-usable local feature variation under small pose and contact-layout perturbations, not a calibrated scene model.
Diffusion features (DIFT)~\cite{zhang2023tale} provide a promising source of this property: their denoising hierarchy preserves dense correspondence cues while retaining semantic context.

Existing representation routes only partially address this property.
Robotics-specific and generalist encoders such as R3M~\cite{nair2022r3m}, VIP~\cite{ma2022vip}, and OpenVLA~\cite{kim2024openvla} improve task relevance or instruction grounding~\cite{team2024octo, brohan2022rt, black2410pi0}, but still emphasize task-level semantics over dense local correspondence.
Point-cloud policies~\cite{ze20243d, jia2025pointmappolicy, donat2025towards, haldar2025point} provide explicit 3D structure, yet depend on depth, calibration, and robust sensing.
Geometry-oriented foundation models such as VGGT~\cite{wang2025vggt} and FastVGGT~\cite{shen2026fastvggt} recover metric or multi-view structure from RGB, offering complementary spatial priors, but reconstruction-oriented geometry is not a \emph{policy-usable} feature that discriminates small pose and contact-layout changes.
Conversely, generative diffusion models retain both high-level semantics and fine spatial detail across a multi-scale hierarchy~\cite{rombach2022high,song2020score,dhariwal2021diffusion}: recent analyses show that their U-Net activations encode dense pixel-level correspondences and part--whole structure~\cite{zhang2023tale, stracke2025cleandift}.
However, using diffusion inference directly for online control faces three challenges:
(1) \emph{stochasticity}---denoising noise can translate into action jitter;
(2) \emph{latency}---multi-step inference is too expensive for high-frequency control; and
(3) \emph{representation drift}---naive robot-domain adaptation can distort the pretrained diffusion manifold and collapse correspondence-rich cues.
Appendix~\ref{sec:related_works} provides an extended discussion of these representation families.

To bridge this gap, we propose \textbf{Robot-DIFT}, a diffusion-derived backbone for closed-loop visuomotor control.
Through \emph{Manifold Distillation}, a frozen, noise-conditioned diffusion Teacher transfers correspondence-rich structure into a clean-input, single-pass Student, while annealed teacher anchoring preserves the pretrained diffusion manifold during robot-domain adaptation.
This removes diffusion inference at deployment, yielding a deterministic real-time backbone.
A \emph{Spatial--Semantic Feature Pyramid Network (S2-FPN)} then fuses coarse-to-fine Student decoder features into policy tokens that expose semantic context and fine spatial variation.
We pretrain the backbone once on the large-scale DROID dataset~\cite{khazatsky2024droid}; the Teacher and alignment heads are then removed, leaving a frozen backbone reusable across downstream tasks.

In summary, we make three contributions:
(i) we identify \emph{correspondence sensitivity} as a key representational property for contact-rich manipulation, showing that vision--language and self-supervised objectives tend to underexpose this property whereas diffusion/DIFT features preserve it as dense correspondence;
(ii) we introduce and release \textbf{Robot-DIFT}, a DROID-pretrained, reusable visual backbone that turns this diffusion correspondence prior into policy-usable features through \emph{Manifold Distillation} and \emph{S2-FPN}; the resulting backbone is deterministic, single-pass, real-time, and frozen for reuse across downstream robot policies; and
(iii) we provide an evidence chain across mechanism analysis, controlled ablations, RoboCasa, LIBERO-10, and real-robot experiments, showing that Robot-DIFT improves contact-sensitive manipulation over vision--language, self-supervised, geometry-oriented, and diffusion baselines, and that S2-FPN \emph{unlocks} rather than replaces the diffusion correspondence prior.

%% file: sections/3_method.tex
\section{Method}
\label{sec:method}
\begin{figure}[t]
\centering
\includegraphics[width=\textwidth]{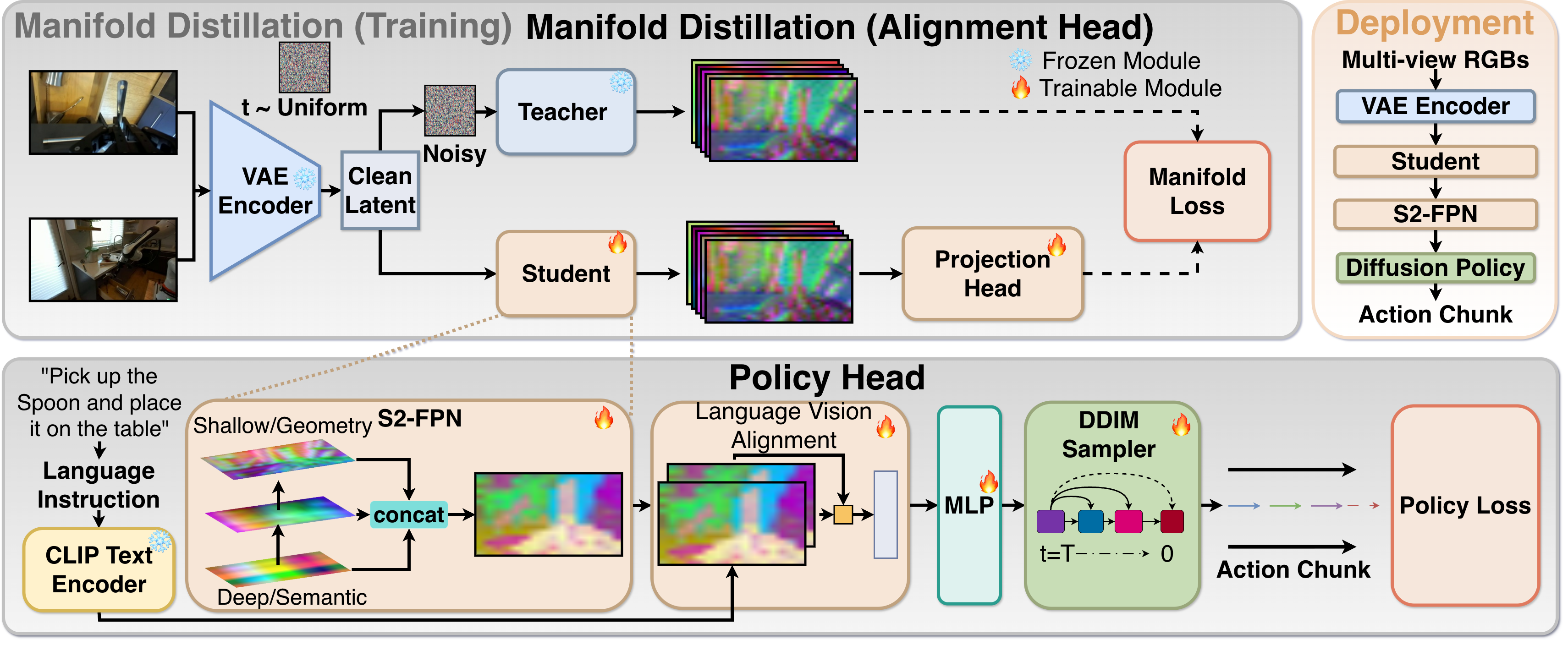}
\caption{\textbf{Robot-DIFT pipeline.}
During DROID pretraining, Manifold Distillation trains a clean-input Student under a frozen, noise-conditioned diffusion Teacher, using alignment heads to match the correspondence-rich diffusion feature manifold; S2-FPN fuses the multi-scale Student features into the policy's visual tokens. After pretraining, the Teacher and alignment heads are removed, leaving a frozen, deterministic, single-pass backbone.
}
\label{fig:pipeline}
\vspace{-1.35em}
\end{figure}

Robot-DIFT adapts correspondence-rich diffusion features into a deterministic visual backbone for robotic control (Fig.~\ref{fig:pipeline}).
It is pretrained once on DROID and then reused as a frozen backbone for downstream manipulation policies, and has two components.
\emph{Manifold Distillation} (Sec.~\ref{sec:manifold_distillation}) trains a clean-input Student U-Net to match a frozen, noise-conditioned diffusion Teacher through alignment heads, preserving the Teacher feature manifold during DROID pretraining.
A \emph{Spatial--Semantic Feature Pyramid Network (S2-FPN)} (Sec.~\ref{sec:s2_fpn}) then fuses multi-scale Student decoder features into the visual tokens consumed by the policy, exposing both semantic context and fine contact-relevant spatial variation.
Appendix~\ref{app:preliminaries} provides latent-diffusion background and feature definitions.

\vspace{-0.6em}
\subsection{Manifold Distillation}
\label{sec:manifold_distillation}
Manifold Distillation is a training-only procedure that transfers the correspondence-rich feature manifold of a pretrained diffusion model into a deterministic Student.
We adopt the CleanDIFT-style mechanism~\cite{stracke2025cleandift}, in which a clean-input Student matches a frozen, noise-conditioned diffusion Teacher through timestep-conditioned projection heads, producing policy-usable features from clean latents in a single forward pass.
The key difference is \emph{what the alignment is for}: CleanDIFT distills task-agnostic descriptors offline, with alignment as the training goal, whereas Robot-DIFT couples the same alignment with an imitation-policy objective during DROID pretraining, using it as an anti-drift regularizer that preserves diffusion correspondences while the backbone adapts to the robot domain (policy-aware annealing, below).
At deployment, the Teacher and alignment heads are removed, so inference stays deterministic without iterative diffusion sampling.

\noindent\textbf{Teacher--Student setup.}
We instantiate a frozen diffusion Teacher U-Net $f_{\theta_0}$ from Stable Diffusion v2.1 and a Student $f_{\theta}$ with an identical U-Net backbone.
The Student is initialized by weight copying, $\theta \leftarrow \theta_0$, while additional modules such as S2-FPN fusion blocks, projection heads, and policy adapters are randomly initialized.
During training, the Teacher remains fixed; only the Student-side parameters are updated.

\noindent\textbf{Noise-conditioned manifold supervision with single-timestep sampling.}
Robotic deployment requires deterministic, single-pass inference on clean latents $\mathbf{z}_0$, whereas the diffusion Teacher naturally defines its feature manifold under noise-conditioned latents $\mathbf{z}_{\tau}$.
We therefore align Student features computed from $\mathbf{z}_0$ to Teacher features computed from $\mathbf{z}_{\tau}$.
Stable Diffusion v2.1 uses a 1000-step diffusion schedule, and each timestep corresponds to a different noise level on the diffusion manifold.
For each training sample and camera view, we draw $\tau$ uniformly from $\{1,\ldots,999\}$ and construct
\begin{equation}
\mathbf{z}_{\tau}
=
\sqrt{\bar{\alpha}_{\tau}}\,\mathbf{z}_0
+
\sqrt{1-\bar{\alpha}_{\tau}}\,\boldsymbol{\epsilon},
\quad
\boldsymbol{\epsilon}\sim\mathcal{N}(\mathbf{0},\mathbf{I}).
\end{equation}

We sample one timestep per view per iteration to keep Teacher supervision efficient in multi-camera training.
Since Teacher feature extraction dominates the computational cost, sampling multiple timesteps would require repeated Teacher forward passes and scale training cost linearly with the number of timesteps.
In our setting, different camera views and training iterations already provide stochastic diversity, making a single $\tau$ per view sufficient in practice while substantially reducing overhead.

\noindent\textbf{Layer-wise alignment across the decoder hierarchy.}
Let $\mathbf{f}^{(k)}_{\tau}$ be the activation from the $k$-th Teacher decoder block under $(\mathbf{z}_{\tau},\tau)$, and let $\mathbf{s}^{(k)}$ be the corresponding Student activation under $\mathbf{z}_0$, with $k\in\{1,\ldots,K\}$.
Timestep-conditioned projection heads $\{g_{\phi}^{(k)}\}_{k=1}^{K}$ map Student features into the Teacher feature space.
The timestep conditioning is confined to these projection heads, so the Student backbone itself remains timestep-free and deterministic at deployment.
The Teacher activations serve as fixed anchors: they are not projected, and gradients are stopped on the Teacher branch.
Thus, the alignment loss constrains only the Student side and mitigates representation drift induced by robot-domain adaptation and policy optimization.

We $\ell_2$-normalize the feature vectors for stability:
\begin{equation}
    \mathbf{u}^{(k)}_{\tau} =
    \frac{g_{\phi}^{(k)}(\mathbf{s}^{(k)}, \tau)}
    {\|g_{\phi}^{(k)}(\mathbf{s}^{(k)}, \tau)\|_2},\quad
    \mathbf{v}^{(k)}_{\tau} =
    \frac{\mathbf{f}^{(k)}_{\tau}}
    {\|\mathbf{f}^{(k)}_{\tau}\|_2},
\end{equation}
where $\mathbf{s}^{(k)}$ are Student features and $\mathbf{f}^{(k)}_{\tau}$ are the frozen Teacher features.
The alignment loss minimizes the negative cosine similarity:
\begin{equation}
\label{eq:align}
\mathcal{L}_{\text{align}}
=
\mathbb{E}_{x\sim\mathcal{D},\,\tau\sim\mathcal{T}}
\left[
    \sum_{k=1}^{K} w_k
    \left(1 - \left\langle \mathbf{u}^{(k)}_{\tau},\, \mathbf{v}^{(k)}_{\tau} \right\rangle \right)
\right],
\end{equation}
where $\mathcal{D}$ is the training dataset, $\mathcal{T}$ is the uniform distribution over $\{1,\ldots,999\}$, and we use uniform weights ($w_k{=}1$) for all blocks.

\noindent\textbf{Policy-aware annealing during DROID pretraining.}
During DROID pretraining, we use $\mathcal{L}_{\text{align}}$ to stabilize the Student while it is optimized with the imitation-policy objective.
Early in pretraining, policy gradients from the randomly initialized readout and action head can be non-stationary and may pull the Student away from the pretrained diffusion manifold.
We therefore start with a stronger Teacher anchor.
As training progresses, a strong alignment constraint can over-regularize the backbone and limit robot-domain adaptation.
We anneal the alignment weight to balance manifold preservation and task-relevant plasticity:
\begin{equation}
\mathcal{L}
=
\mathcal{L}_{\text{policy}}
+
\lambda(t)\,\mathcal{L}_{\text{align}},
\qquad
\lambda(t)=
\lambda_{\min}
+
(\lambda_0-\lambda_{\min})\cdot
\max\!\left(0,\,1-\frac{t}{T_{\text{decay}}}\right).
\end{equation}
Here $t$ denotes the training step.
We set $\lambda_0{=}0.1$, $\lambda_{\min}{=}0.001$, and $T_{\text{decay}}=\lfloor 0.5\,T\rfloor$, where $T$ is the total number of training steps.
For more details regarding the optimization protocol and training hyperparameters, see Appendix~\ref{app:implementation}.

\subsection{Student Architecture: S2-FPN}
\label{sec:s2_fpn}

After Manifold Distillation, the Student processes clean latents $\mathbf{z}_0$ in one forward pass, avoiding the cost and stochastic variance of iterative denoising.
The distilled Student already carries correspondence-rich structure across its decoder hierarchy; S2-FPN exposes this structure to the policy by fusing features across scales, giving access to both \textit{what} to manipulate (semantics) and \textit{where} to interact (contact-relevant spatial structure).

\noindent\textbf{Multi-Scale Feature Extraction.}
The SD2.1 U-Net decoder exposes a full hierarchy of feature maps (Appendix~\ref{app:preliminaries} details the latent-diffusion backbone and the available decoder layers).
Given an image $x$, we extract three of these maps from the Student decoder:
\begin{equation}
\mathcal{S}=\{\mathbf{s}^{(i)}\}_{i=1}^3=\{\mathbf{s}^{(us3)},\mathbf{s}^{(us6)},\mathbf{s}^{(us8)}\},
\end{equation}
ordered from coarse to fine resolution; here $\mathrm{us}3,\mathrm{us}6,\mathrm{us}8$ denote the outputs of the 3rd, 6th, and 8th upsampling (decoder) blocks of the Student U-Net.
We choose these levels to span the semantic--spatial trade-off:
$\mathbf{s}^{(us3)}$ provides global context,
$\mathbf{s}^{(us6)}$ captures affordance structure,
and $\mathbf{s}^{(us8)}$ preserves local edges critical for precise contact.
We analyze alternative layer selections in Sec.~\ref{sec:ablation}.

\noindent\textbf{Global-to-Fine Feature Fusion.}
Direct concatenation can be brittle because feature maps differ in scale and semantic level.
Global-to-Fine Fusion instead injects global semantic context into high-resolution spatial maps (Fig.~\ref{fig:pipeline}).
Let $\mathbf{s}^{(i)}$ be the raw feature at pyramid level $i$ (with $i\in\{1,2,3\}$ corresponding to $\{\mathbf{s}^{(us3)},\mathbf{s}^{(us6)},\mathbf{s}^{(us8)}\}$, ordered coarse to fine), and let $\mathbf{M}^{(i)}$ be the fused map.
We initialize $\mathbf{M}^{(1)}=\mathbf{s}^{(1)}$ from the coarsest semantic feature and compute:
\begin{equation}
    \mathbf{M}^{(i)} = \operatorname{ConvBlock}\!\left(\mathbf{s}^{(i)} \oplus \operatorname{Upsample}\!\left(\mathbf{M}^{(i-1)}\right)\right),\qquad i\in\{2,3\}.
\end{equation}
Here $\oplus$ is channel-wise concatenation, and $\operatorname{ConvBlock}$ is a residual stack of convolution, group normalization, and GELU.
$\operatorname{Upsample}(\cdot)$ uses bilinear interpolation to match the spatial resolution of $\mathbf{s}^{(i)}$.
The resulting maps are spatially aligned, preserve fine contact-relevant detail, and remain informed by global semantics.
Because S2-FPN mirrors the spatial resolutions and channel structure of the Stable Diffusion U-Net decoder, it also supports layer-wise alignment in Manifold Distillation.
By propagating global semantics before policy pooling, S2-FPN reduces cross-scale mismatch and surfaces the backbone's correspondence sensitivity to the policy without sacrificing semantic coherence.



\noindent\textbf{Policy interface.}
The fused visual map is converted into a fixed-dimensional observation embedding through a lightweight language-conditioned readout.
We flatten the visual map into spatial tokens and retain image coordinates with positional encodings.
A frozen CLIP text encoder provides instruction tokens that condition the visual readout through a small cross-attention module.
For multi-view inputs, the readout is applied independently to each view, and the resulting tokens are aggregated before being passed to the policy head.
This exposes S2-FPN features to language-conditioned control.
Implementation details are provided in Appendix~\ref{app:lang-vis-align}.

\noindent\textbf{Deployment.}
After pretraining, we remove the Teacher and projection heads $\{g_{\phi}^{(k)}\}_{k=1}^{K}$, keeping only the distilled Student and S2-FPN as the frozen Robot-DIFT backbone.
Downstream policies reuse this backbone for deterministic single-pass features that preserve diffusion-derived correspondence sensitivity without iterative diffusion inference.

%% file: sections/4_experiment.tex
\vspace{-0.6em}
\section{Experiments}
\label{sec:experiments}

We organize the experiments around four research questions.
\textbf{RQ1: Correspondence Sensitivity and Control.}
Do Robot-DIFT features preserve correspondence sensitivity, and does this improve contact-sensitive control?
\textbf{RQ2: Policy-Facing Diffusion Features.}
Which components make diffusion-derived features usable by a downstream policy?
\textbf{RQ3: Language-Conditioned Transfer.}
Can a frozen DROID-pretrained backbone transfer to language-conditioned manipulation in real time?
\textbf{RQ4: Real-Robot Contact Validation.}
Does the same contact-sensitive profile appear on physical robots?

To isolate the effect of the visual representation or policy-facing readout, we hold the policy class, demonstrations, training protocol, and metrics constant throughout our controlled comparisons.
Implementation details are provided in Appendix~\ref{app:experimental_details} and Appendix~\ref{app:empirical_analysis}.

\vspace{-0.6em}
\subsection{Correspondence Sensitivity and Control} \label{sec:sim_benchmarks}
We answer RQ1 with two tests: a feature-level diagnostic of local discrimination within the same object, and a RoboCasa~\cite{nasiriany2024robocasa} benchmark that fixes the policy class and training protocol while varying the visual representation.

\begin{figure*}[t]
\centering
\includegraphics[width=\textwidth]{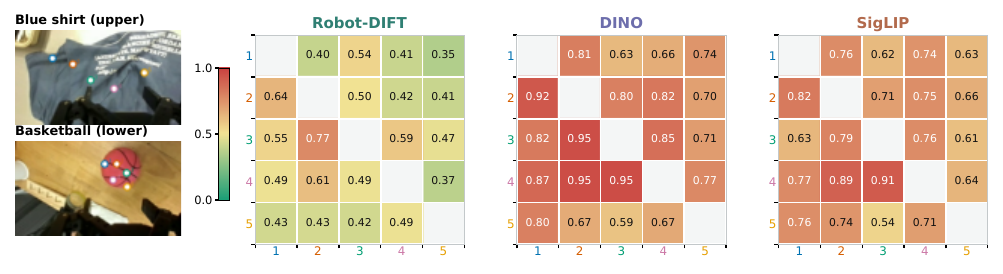}
\caption{\textbf{Object-centric local feature sensitivity.}
Pairwise cosine similarity among five same-object points across two scenes.
Lower is more locally discriminative.
Robot-DIFT has lower mean intra-object similarity (0.48) than DINOv3 (0.78) and SigLIP (0.72).}
\label{fig:objectcentric_correlation}
\vspace{-1.35em}
\end{figure*}

\noindent\textbf{Object-centric local sensitivity.}
This diagnostic tests whether a visual feature can distinguish different surface points on the same manipulated object.
For each image, we sample five object-surface points and compute the pairwise cosine-similarity matrix between their features (Fig.~\ref{fig:objectcentric_correlation}).
Since all points share one object, high similarity reflects object identity, while lower similarity means local surface regions remain distinguishable.
Robot-DIFT shows lower point-to-point similarity than DINOv3 and SigLIP, indicating that it preserves more local visual information for manipulation.
We next test whether this difference improves control in RoboCasa under the same policy and training protocol.


\noindent\textbf{RoboCasa protocol.}
We then evaluate control on the 24-task \emph{RoboCasa} subset, grouped into seven task categories.
All rows use the same Diffusion Policy~\cite{chi2025diffusion} recipe, and success is averaged over 50 rollouts per task.
The comparison covers vision--language, geometry-oriented, self-supervised, generative, and DROID-adapted visual representations (Table~\ref{tab:robocasa-benchmark}).
Full per-task results, baseline definitions, and protocol details are in Appendix Table~\ref{tab:robocasa-benchmark-full} and Appendix~\ref{app:sim_setup}.
Appendix Fig.~\ref{fig:corl_representation_evidence} reports the corresponding latency profile.

\begin{table*}[!t]
\centering
\caption{\textbf{Representation choice and deployment cost on RoboCasa.}
Success values are category means over 50 rollouts per task.
The final row reports the inference times for different encoders.}
\label{tab:robocasa-benchmark}
\scriptsize
\setlength{\tabcolsep}{3.0pt}
\renewcommand{\arraystretch}{1.08}
\resizebox{\textwidth}{!}{
\begin{tabular}{lcccccccccc}
\toprule
\multirow{2}{*}{\textbf{Category}} &
\multicolumn{2}{c}{\textbf{Vision--language}} &
\multicolumn{2}{c}{\textbf{Geometry-oriented}} &
\multicolumn{2}{c}{\textbf{Self-supervised}} &
\multicolumn{2}{c}{\textbf{Generative}} &
\multicolumn{2}{c}{\textbf{DROID-adapted}} \\
\cmidrule(lr){2-3}\cmidrule(lr){4-5}\cmidrule(lr){6-7}\cmidrule(lr){8-9}\cmidrule(lr){10-11}
& \makecell[c]{\textbf{CLIP}\\\cite{radford2021learning}} & \makecell[c]{\textbf{SigLIP}\\\cite{zhai2023sigmoid}} & \makecell[c]{\textbf{VGGT}\\\cite{wang2025vggt}} & \makecell[c]{\textbf{FastVGGT}\\\cite{shen2026fastvggt}} & \makecell[c]{\textbf{DINOv2}\\\cite{oquab2023dinov2}} & \makecell[c]{\textbf{DINOv3}\\\cite{simeoni2025dinov3}} & \makecell[c]{\textbf{SiT}\\\cite{ma2024sit}} & \makecell[c]{\textbf{DIFT}\\\cite{zhang2023tale}} & \makecell[c]{\textbf{Robot-DINO}} & \makecell[c]{\textbf{Robot-DIFT}\\\textbf{(Ours)}} \\
\midrule
Pick-and-place & 0.03 & 0.02 & 0.04 & 0.05 & 0.03 & 0.03 & 0.07 & 0.06 & 0.10 & \textbf{0.12} \\
Doors & 0.23 & 0.42 & 0.43 & 0.46 & 0.45 & 0.45 & 0.49 & 0.49 & 0.51 & \textbf{0.73} \\
Drawers & 0.51 & 0.61 & 0.62 & 0.62 & 0.66 & 0.78 & 0.54 & \textbf{0.85} & 0.82 & 0.80 \\
Knobs & 0.18 & 0.17 & 0.16 & 0.14 & 0.21 & 0.33 & 0.15 & \textbf{0.40} & 0.34 & 0.33 \\
Levers & 0.35 & 0.49 & 0.43 & 0.44 & 0.45 & 0.51 & 0.35 & 0.58 & 0.56 & \textbf{0.61} \\
Buttons & 0.29 & 0.47 & 0.15 & 0.16 & 0.48 & 0.61 & 0.58 & 0.69 & 0.76 & \textbf{0.91} \\
Insertion & 0.06 & 0.06 & 0.00 & 0.00 & 0.08 & 0.17 & 0.15 & 0.20 & 0.20 & \textbf{0.46} \\
\midrule
\textbf{Average} & 0.19 & 0.27 & 0.22 & 0.23 & 0.28 & 0.33 & 0.29 & 0.38 & 0.40 & \textbf{0.49} \\
\textbf{Backbone latency (ms)} & \textbf{19} & 59 & 288 & 85 & 30 & 28 & 77 & 66 & 31 & 68 \\
\bottomrule
\end{tabular}
}
\vspace{-1.35em}
\end{table*}

\noindent\textbf{Results and Analysis.}
Robot-DIFT achieves the best average success, reaching 0.49 across the RoboCasa subset.
The gains are strongest on tasks that require precise local alignment, especially Buttons and Insertion, while raw DIFT remains competitive on Drawers and Knobs.
Robot-DINO provides an important control.
Although it reaches a competitive average (0.40), its gains concentrate on coarser tasks, and it stays well behind Robot-DIFT on the most contact-sensitive ones (Buttons 0.76 vs.\ 0.91, Insertion 0.20 vs.\ 0.46).
This suggests that robot-domain adaptation alone does not create the diffusion-style correspondence sensitivity needed for precise contact.
Moreover, directly finetuning DINO without manifold preservation can distort its pretrained features and performs worse, as shown in Appendix Fig.~\ref{fig:align_ablation_coffee}.
We also compare two generative backbones, SiT and DIFT, and find that the transformer-based SiT lags behind the U-Net-based DIFT (0.29 vs.\ 0.38), likely because its tokenized features lack the 2D spatial structure that convolutional diffusion features retain for dense correspondence.
VGGT and FastVGGT provide static-scene geometry priors, but their multi-view reconstruction objective is mismatched to dynamic closed-loop manipulation, which may explain why they remain lower under the same policy interface.
Overall, the results indicate that diffusion-derived correspondence cues help contact-sensitive control, but direct DIFT extraction is not sufficient; robot-domain adaptation and a policy-facing readout are both needed.

\begin{table*}[!t]
\centering

\caption{\textbf{Robot-DIFT representation controls.}
(a) Contact-sensitive RoboCasa controls isolating multi-scale feature extraction and annealed manifold alignment.
(b) LIBERO-10 policy-interface controls comparing single-layer, concatenation, and S2-FPN readouts on original, non-DROID-adapted visual backbones.}
\label{tab:robotdift_ablations}
\footnotesize
\renewcommand{\arraystretch}{1.06}
\setlength{\tabcolsep}{3.4pt}

\begin{minipage}[t]{0.42\textwidth}
\centering
\begin{tabular}{@{}lccc@{}}
\toprule
\textbf{Variant} & \textbf{Btn.} & \textbf{Ins.} & \textbf{Avg. $\uparrow$} \\
\midrule
\multicolumn{4}{@{}l}{\textit{Feature extraction}} \\
\hspace{1mm} SingleScale-us3 & 0.39 & 0.13 & 0.26 \\
\hspace{1mm} SingleScale-us6 & 0.52 & 0.28 & 0.40 \\
\hspace{1mm} SingleScale-us8 & 0.46 & 0.24 & 0.35 \\
\hspace{1mm} \textbf{MultiScale (Ours)} & \textbf{0.90} & \textbf{0.46} & \textbf{0.68} \\
\midrule
\multicolumn{4}{@{}l}{\textit{Alignment schedule}} \\
\hspace{1mm} NoAnneal & 0.76 & 0.31 & 0.54 \\
\hspace{1mm} \textbf{Anneal (Ours)} & \textbf{0.90} & \textbf{0.46} & \textbf{0.68} \\
\bottomrule
\end{tabular}
\end{minipage}
\hspace{0.035\textwidth}
\begin{minipage}[t]{0.50\textwidth}
\centering
\begin{tabular}{@{}lllc@{}}
\toprule
\textbf{Backbone} & \textbf{Readout} & \textbf{Layers} & \textbf{Success $\uparrow$} \\
\midrule
DINOv2 & Single & default & 0.58 \\
DINOv2 & Concat & L2/L5/L8/L11 & 0.78 \\
DINOv2 & S2-FPN & L2/L5/L8/L11 & \textbf{0.83} \\
\midrule
DIFT & Single & us6 & 0.76 \\
DIFT & Concat & us3/us6/us8 & 0.88 \\
DIFT & S2-FPN & us3/us6/us8 & \textbf{0.91} \\
\bottomrule
\end{tabular}
\end{minipage}

\vspace{-1.2em}
\end{table*}

\vspace{-0.6em}
\subsection{Representation Interface and Distillation Controls}
\label{sec:ablation}

RQ2 asks which components make diffusion-derived features usable by a downstream policy.
We test two complementary aspects: distillation-side design choices on two representative contact-sensitive RoboCasa tasks, \texttt{PressButton}~(Btn) and \texttt{Insertion}~(Ins), and policy-facing readouts on LIBERO-10 using original, non-DROID-adapted visual backbones.
Table~\ref{tab:robotdift_ablations}(a) reports feature-scale and alignment controls, while Table~\ref{tab:robotdift_ablations}(b) reports readout controls for DINOv2 and DIFT.

\noindent\textbf{Distillation-side controls.}
Table~\ref{tab:robotdift_ablations}(a) shows that multi-scale Student features are important for contact-rich control.
MultiScale improves average success from 0.40 for the best single-scale variant to 0.68.
Annealed alignment further improves over NoAnneal from 0.54 to 0.68.
These results support the two roles of Manifold Distillation: preserving the diffusion feature manifold early in training and allowing robot-domain adaptation later.

\noindent\textbf{Policy-interface controls.}
Table~\ref{tab:robotdift_ablations}(b) evaluates policy-facing readouts on LIBERO-10.
All visual backbones in this control are original, off-the-shelf backbones and are not adapted on DROID.
For both DINOv2 and DIFT, multi-scale readout improves downstream policy performance over a single-layer readout.
S2-FPN further outperforms direct concatenation for both backbones, indicating that structured multi-scale fusion is generally useful when a backbone provides features at different levels.
At the same time, the best result is obtained by DIFT with S2-FPN, showing that the readout improves feature usage but does not replace the diffusion-derived backbone prior.

Together, these controls show that Robot-DIFT depends on both a stable diffusion-derived feature manifold and a policy-facing interface that exposes multi-scale visual structure.
Additional finetuning evidence and per-task RoboCasa results are provided in Appendix Fig.~\ref{fig:align_ablation_coffee} and Appendix Table~\ref{tab:robocasa-benchmark-full}.

\vspace{-0.6em}
\subsection{Language-Conditioned Transfer}
\label{sec:libero_transfer}

RQ3 asks whether the DROID-pretrained Robot-DIFT backbone transfers beyond RoboCasa.
Unlike the readout controls in Sec.~\ref{sec:ablation}, which use off-the-shelf backbones, here we deploy the full DROID-pretrained Robot-DIFT backbone.
We evaluate this on \emph{LIBERO-10} under the UVA protocol~\cite{li2025unified}, comparing against action-only imitation policies, VLA models, and video-action systems.
We report average task success and trajectory inference time.

\noindent\textbf{Results and Analysis.}
Table~\ref{tab:transfer_realrobot}(a) shows that Robot-DIFT achieves the strongest accuracy--latency trade-off.
With a frozen backbone and a standard imitation-policy head, it reaches 0.93 success with 0.01s trajectory inference, improving over UVA in both success and speed and over $\pi_0$ in success while being faster.
Robot-DIFT can also replace the visual representation inside OpenVLA; we call this variant \emph{DIFT-VLA}.
After lightweight fine-tuning, DIFT-VLA improves success from 0.54 to 0.68 over the original OpenVLA.
These results show that Robot-DIFT transfers to language-conditioned manipulation as both an efficient imitation-policy backbone and a reusable visual representation for VLA pipelines.



\begin{table*}[!t]
\centering
\caption{\textbf{Transfer and physical validation.}
(a) LIBERO-10 results comparing Robot-DIFT with action-only, VLA, and video-action systems; speed denotes trajectory inference time, measured over a 16-step predicted trajectory with 8 executed actions, while OpenVLA is run for 8 single-action forward passes to match the executed horizon.
(b) Real-robot validation across coarse and contact-sensitive manipulation tasks.}

\label{tab:transfer_realrobot}
\scriptsize
\renewcommand{\arraystretch}{1.06}

\begin{minipage}[t]{0.53\textwidth}
\centering
\setlength{\tabcolsep}{2.8pt}
\begin{tabular}{@{}llcc@{}}
    \toprule
    \textbf{Family} & \textbf{Method} & \textbf{Success $\uparrow$} & \textbf{Latency (s) $\downarrow$} \\
    \midrule
    \multirow{2}{*}{Act.-only}
    & DP-C~\cite{chi2025diffusion} & 0.53 & 0.50 \\
    & DP-T~\cite{chi2025diffusion} & 0.58 & 0.36 \\
    \midrule
    \multirow{3}{*}{VLA}
    & OpenVLA~\cite{kim2024openvla} & 0.54 & 1.52 \\
    & $\pi_0$~\cite{black2410pi0} & 0.85 & 0.09 \\
    & $\pi_0$-FAST~\cite{black2410pi0} & 0.60 & 0.09 \\
    \midrule
    \multirow{2}{*}{Vid.-act.}
    & UniPi~\cite{du2023learning} & 0.00 & 24.07 \\
    & UVA~\cite{li2025unified} & 0.90 & 0.23 \\
    \midrule
    OpenVLA+rep. & DIFT-VLA & 0.68 & 1.75 \\
    DP+rep. & \textbf{Robot-DIFT (Ours)} & \textbf{0.93} & \textbf{0.01} \\
    \bottomrule
\end{tabular}
\end{minipage}
\hspace{0.035\textwidth}
\begin{minipage}[t]{0.40\textwidth}
\centering
\setlength{\tabcolsep}{1.3pt}
\begin{tabular}{@{}lccccc@{}}
    \toprule
    \multirow{2}{*}{\textbf{Backbone}} &
    \multicolumn{2}{c}{\textbf{Coarse/artic.}} &
    \multicolumn{2}{c}{\textbf{Contact}} &
    \multirow{2}{*}{\textbf{Avg. $\uparrow$}} \\
    \cmidrule(lr){2-3}\cmidrule(lr){4-5}
    & Sort & Open & Insert & Press & \\
    \midrule
    DINOv2~\cite{oquab2023dinov2} & \textbf{0.55} & \textbf{0.60} & 0.05 & 0.35 & 0.39 \\
    SigLIP~\cite{zhai2023sigmoid} & 0.40 & 0.45 & 0.00 & 0.10 & 0.24 \\
    \textbf{Robot-DIFT (Ours)} & 0.55 & 0.55 & \textbf{0.35} & \textbf{0.55} & \textbf{0.50} \\
    \bottomrule
\end{tabular}
\end{minipage}

\vspace{-1.4em}
\end{table*}
\vspace{-0.6em}
\subsection{Real-Robot Experiments}
\label{sec:real_robot}
RQ4 tests whether the contact-sensitive profile transfers to physical manipulation.
We evaluate four rigid-object tasks that range from coarse semantic interaction to tight contact: \emph{Sort Cup}, \emph{Open Lid}, \emph{Insert Pin}, and \emph{Press Switch}.
All methods use the same robot, cameras, demonstrations, policy class, and frozen visual backbones.
Only the policy-facing representation changes.
DINOv2 and SigLIP serve as semantic baselines.
Each task is trained on roughly 25--30 demonstration trajectories.
Because absolute success depends on data and task difficulty, our claims concern the \emph{relative} difference between representations under this fixed protocol, not the absolute values.
Task images, hardware details, and evaluation protocols are provided in Appendix~\ref{app:real_robot_details} and Fig.~\ref{fig:real_robot_task}.

\noindent\textbf{Results and Analysis.}
Table~\ref{tab:transfer_realrobot}(b) shows that DINOv2 remains competitive on \emph{Sort Cup} and \emph{Open Lid}, where final alignment is less constrained than in insertion or switch pressing.
Robot-DIFT improves most on contact-sensitive tasks.
\emph{Insert Pin} increases from 0.05 with DINOv2 to 0.35, and \emph{Press Switch} increases from 0.35 to 0.55.
The gains appear where local pre-contact alignment and precise contact placement matter most.
This matches the contact-sensitive profile observed in RoboCasa and the ablations in Table~\ref{tab:robotdift_ablations}(a).
This intentionally controlled suite provides evidence for visual pre-contact alignment, not a solution to terminal contact physics, force feedback, deformable objects, or broader embodiment variation.
\vspace{-0.6em}

%% file: sections/5_conclusion.tex
\section{Conclusion and Limitations}
\label{sec:conclusion}

We introduced Robot-DIFT, a diffusion-derived visual backbone for closed-loop manipulation, showing that contact-rich control needs correspondence-sensitive features beyond semantic recognition. Robot-DIFT makes this prior practical by distilling a noise-conditioned teacher into a deterministic Student, constraining feature drift during DROID pretraining, and exposing multi-scale features through S2-FPN. Across RoboCasa, LIBERO-10, and real-robot tasks, it improves contact-sensitive manipulation, transfers to language-conditioned control, and serves as a reusable interface for imitation and VLA-style policies---complementary to, not a replacement for, generalist VLA training. Limitations include fixed real-robot settings, rigid-object tasks, no force or tactile feedback, reliance on an SD2.1 teacher, and offline DROID pretraining cost.

%% file: sections/appendix.tex
\section{Appendix Roadmap}
\label{app:appendix_roadmap}

This appendix follows the evidence ladder in Sec.~\ref{sec:experiments}.
Appendix~\ref{app:representation_interface_motivation} expands the representation-interface motivation compressed in the Introduction.
Appendix~\ref{sec:related_works} then situates this motivation in prior work.
Appendix~\ref{app:preliminaries} and Appendix~\ref{app:implementation} give the method and training details behind Manifold Distillation and S2-FPN.
Appendix~\ref{app:sim_setup} and Appendix~\ref{app:real_robot_details} specify the simulation and physical evaluation protocols.
Appendix~\ref{app:empirical_analysis} then mirrors the main experiment questions: controlled encoder swaps, readout controls, and Manifold Distillation diagnostics.

\section{Extended Representation-Interface Motivation}
\label{app:representation_interface_motivation}

This section preserves the longer motivation behind the compressed Introduction.
It clarifies how we use \emph{correspondence sensitivity}, why semantic and geometric priors are not sufficient by themselves, and why diffusion features require distillation before deployment.
These points are intended as conceptual grounding for the controlled comparisons in Sec.~\ref{sec:experiments}, not as a separate metric beyond the reported encoder swaps, readout controls, diagnostics, and contact-sensitive manipulation results.

\paragraph{Correspondence sensitivity as a control-facing property.}
Contact-rich policies require visual feedback that changes predictably with the local state variables that change the next action.
We therefore use \emph{correspondence sensitivity} to mean that behaviorally distinct but nearby object configurations induce spatially local, repeatable, and discriminative changes in the policy-facing features.
An operational intuition is that features near a contact-relevant region should respond more reliably to small pose or contact-layout changes than unrelated background features.
In this paper, this property is tested indirectly through controlled representation swaps, readout controls, latent-space diagnostics, and the task profile of contact-sensitive manipulation; we do not introduce a standalone perturbation-response metric in the main evaluation.

\paragraph{Why semantic recognition alone can be insufficient.}
Vision--language and self-supervised recognition encoders provide strong priors for identifying objects and grounding instructions.
However, the invariance that benefits recognition can attenuate small pose offsets, local boundaries, and pre-contact alignments that are not nuisance variables for control.
This creates the failure mode emphasized in the main text: a policy can be visually aware of \emph{what} to interact with, yet weakly informed about \emph{where} and \emph{how} the next correction should be made.
Our experiments therefore compare semantic, self-supervised, geometry-oriented, generative, and robot-domain-adapted representations under the same policy protocol, rather than treating object recognition performance as a proxy for control suitability.

\paragraph{Why metric geometry is complementary but not equivalent.}
Explicit point-cloud or depth-based pipelines provide useful 3D structure, but they can introduce practical and statistical constraints: camera--robot calibration, depth alignment, sensor noise, resolution limits, and reduced standardization across sensors and scenes.
Geometry-oriented foundation models such as VGGT and FastVGGT reduce some of these requirements by predicting metric or multi-view spatial structure from RGB inputs.
We use them as geometry-oriented controls because they test a natural alternative explanation: if metric geometry alone were the missing ingredient, these features should close the gap under the same fixed-encoder policy interface.
The results in Table~\ref{tab:robocasa-benchmark}, Appendix Table~\ref{tab:robocasa-benchmark-full}, and Appendix Fig.~\ref{fig:corl_representation_evidence} show a different pattern: VGGT/FastVGGT provide complementary spatial priors, but their policy-facing features remain weaker on the contact-sensitive subset, and VGGT has the largest deployment cost among the completed RoboCasa rows.

\paragraph{Why diffusion features need a deployable interface.}
Diffusion features are attractive because denoising models preserve a hierarchy of semantic and spatial cues, and prior DIFT/CleanDIFT analyses show strong dense correspondence behavior~\cite{zhang2023tale, stracke2025cleandift}.
Direct diffusion inference, however, is poorly matched to closed-loop control.
Iterative denoising can introduce stochastic variation, requires multiple network evaluations, and makes multi-camera deployment expensive.
Moreover, naively adapting a pretrained visual backbone to robot data can drift away from the original feature distribution.
Robot-DIFT addresses this interface problem by distilling the noise-conditioned diffusion hierarchy into a clean-input Student, anchoring the Student to the Teacher manifold during robot-domain adaptation, and exposing the resulting multi-scale features through S2-FPN.

\input{sections/2_related_works}

\section{Method and Representation Interface Details}
\label{app:preliminaries}

\begin{figure*}[t]
    \centering
    \includegraphics[width=\textwidth]{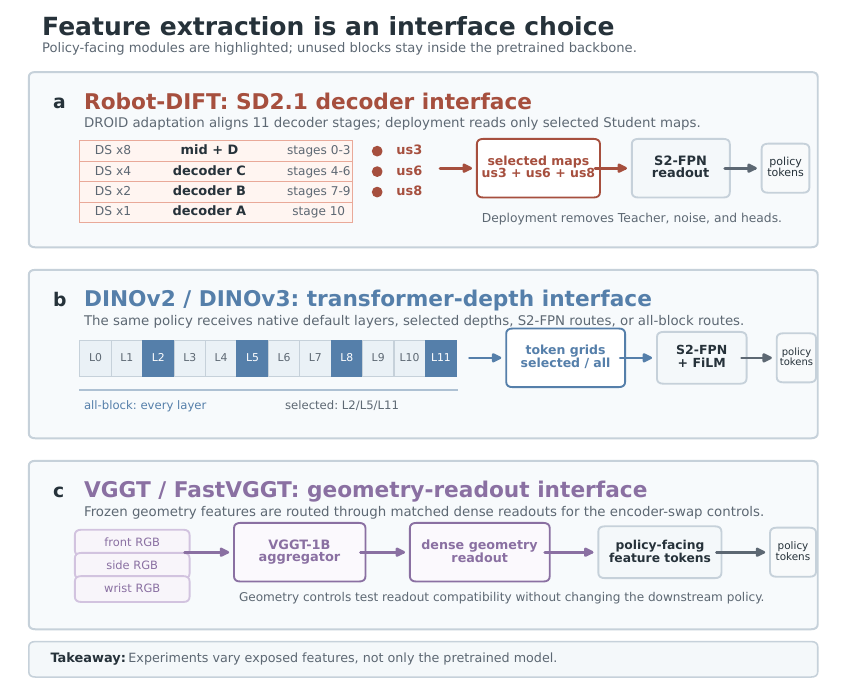}
    \caption{\textbf{Policy-facing feature interfaces used across representation controls.}
    Robot-DIFT reads selected SD2.1 decoder maps through S2-FPN after Teacher--Student alignment on DROID.
    DINO controls expose transformer token grids at matched depths, while VGGT and FastVGGT provide geometry-oriented readouts.
    The comparison fixes each encoder while using the same downstream readout and policy training protocol.}
    \label{fig:feature_alignment}
\end{figure*}

This section expands only the method details needed to interpret the appendix controls: teacher--student feature alignment, S2-FPN fusion, and query pooling.
It also situates Robot-DIFT relative to the DINOv2 readout controls and the VGGT/FastVGGT geometry controls by making explicit where each encoder exposes features to the policy (Fig.~\ref{fig:feature_alignment}).

\subsection{Latent Diffusion Background}
\label{app:ldm_preliminaries}

Robot-DIFT builds on Latent Diffusion Models (LDMs)~\cite{rombach2022high}, specifically Stable Diffusion v2.1.
An LDM uses a pretrained Variational Autoencoder~\cite{kingma2013auto} to map an RGB image $\mathbf{x} \in \mathbb{R}^{H \times W \times 3}$ into a compressed latent representation $\mathbf{z}_0 = \mathcal{E}(\mathbf{x}) \in \mathbb{R}^{h \times w \times c}$.
During diffusion pretraining, Gaussian noise is added to this latent according to a fixed variance schedule:
\begin{equation}
    \mathbf{z}_\tau = \sqrt{\bar{\alpha}_\tau} \mathbf{z}_0 + \sqrt{1 - \bar{\alpha}_\tau} \boldsymbol{\epsilon}, \quad \boldsymbol{\epsilon} \sim \mathcal{N}(\mathbf{0}, \mathbf{I}),
\end{equation}
where $\bar{\alpha}_\tau$ is the noise schedule coefficient.
A time-conditional U-Net $\boldsymbol{\epsilon}_\theta(\mathbf{z}_\tau, \tau, \mathbf{c})$ is trained to denoise $\mathbf{z}_\tau$ conditioned on context $\mathbf{c}$.
Robot-DIFT uses the U-Net's internal decoder activations rather than the generated image.
Prior DIFT analyses show that these decoder features contain dense semantic and spatial correspondences~\cite{zhang2023tale}; Robot-DIFT treats the pretrained SD2.1 U-Net as a frozen Teacher and distills these noise-conditioned features into a clean-input Student for deterministic control.

\subsection{Feature Extraction and Teacher Alignment}
\label{app:diffusion_features}
For Robot-DIFT, we expose hierarchical feature maps from $K=11$ layers of the U-Net decoder (Fig.~\ref{fig:feature_alignment}a). At image coordinate $\mathbf{u}$, point-wise feature vectors $\mathbf{f}^{(k)}_{\tau}(\mathbf{u})$ are obtained by bilinear sampling from the corresponding feature map. At training time, the frozen Teacher receives a noised latent $z_\tau$, while the Student receives the clean latent $z_0$. Following the projection-head alignment idea in CleanDIFT~\cite{stracke2025cleandift}, we attach a timestep-conditioned projection head $g_\phi^{(k)}$ to each Student layer so that clean-input Student features can be compared with noise-conditioned Teacher features.
This broad alignment stage preserves the diffusion feature manifold, whereas the deployed policy reads only the selected Student maps through S2-FPN.
The DINOv2 controls in Fig.~\ref{fig:feature_alignment}b use the same principle at the experiment level: internal transformer features are exposed to the same downstream policy to test which abstraction depths are actionable.
The VGGT/FastVGGT controls in Fig.~\ref{fig:feature_alignment}c provide matched geometry-oriented readouts without turning the paper around a separate geometry model.

For a sampled timestep $\tau$, define
\begin{equation}
\mathbf{u}^{(k)}_{\tau}
=
\frac{g_{\phi}^{(k)}(\mathbf{s}^{(k)},\tau)}
{\|g_{\phi}^{(k)}(\mathbf{s}^{(k)},\tau)\|_2},
\qquad
\mathbf{v}^{(k)}_{\tau}
=
\frac{\operatorname{sg}(\mathbf{f}^{(k)}_{\tau})}
{\|\operatorname{sg}(\mathbf{f}^{(k)}_{\tau})\|_2},
\end{equation}
where $\operatorname{sg}(\cdot)$ stops gradients through the Teacher branch. The alignment objective is
\begin{equation}
\mathcal{L}_{\text{align}}
=
\mathbb{E}_{x,\tau}
\left[
\sum_{k=1}^{K} w_k
\left(1-\left\langle \mathbf{u}^{(k)}_{\tau},\mathbf{v}^{(k)}_{\tau}\right\rangle\right)
\right],
\label{eq:app-align}
\end{equation}
with uniform weights $w_k{=}1$ unless otherwise stated. This formulation anchors a clean-input robot-domain Student to the diffusion-derived spatial feature hierarchy. At inference, $z_\tau$, the Teacher, and all projection heads are removed.

\subsection{S2-FPN Fusion and Aggregation}
\label{app:s2fpn}

To leverage multi-scale spatial and semantic context, S2-FPN uses a global-to-fine FPN-style architecture. Let $\{F_i\}_{i=1}^{L}$ be the selected feature maps sorted from coarse to fine spatial resolution. We first project all maps to a shared dimension $C$:
\begin{equation}
P_i = L_i(F_i),
\qquad i=1,\ldots,L,
\end{equation}
where $L_i$ is a $1\times1$ lateral projection followed by normalization and GELU activation.

\emph{Global-to-fine fusion:} coarse semantic context is propagated toward fine spatial maps:
\begin{equation}
\mathbf{M}_1=P_1,\qquad
\mathbf{M}_i=\rho_i\!\left(
\operatorname{Concat}\!\left[
\operatorname{Up}(\mathbf{M}_{i-1}),P_i
\right]\right),
\quad i=2,\ldots,L .
\label{eq:s2fpn_fusion}
\end{equation}
Here $\rho_i$ denotes a convolution--normalization--GELU fusion block, and $\operatorname{Up}(\cdot)$ uses bilinear interpolation to match the next feature resolution. The final visual map is produced at the finest resolution:
\begin{equation}
\mathbf{M}_{v}
=
\rho_{\mathrm{out}}
\left(\mathbf{M}_L\right).
\label{eq:s2fpn_output}
\end{equation}

\emph{Query aggregation:} We flatten $\mathbf{M}_{v}\in\mathbb{R}^{B\times C_v\times H\times W}$ into spatial tokens $\mathbf{X}_{v}\in\mathbb{R}^{B\times HW\times C_v}$ and add 2D RoPE positional encodings~\cite{su2024roformer,heo2024rotary}. For language-conditioned tasks, frozen CLIP text tokens $\mathbf{X}_{\ell}\in\mathbb{R}^{B\times N_{\ell}\times C_{\ell}}$ are used as queries after a lightweight projection to the shared dimension $C$:
\begin{equation}
\mathbf{H}
=
\operatorname{MHA}
\left(
\operatorname{LN}(\psi_\ell(\mathbf{X}_{\ell})),
\operatorname{LN}(\psi_v(\mathbf{X}_{v})),
\operatorname{LN}(\psi_v(\mathbf{X}_{v}))
\right),
\label{eq:query_pooling}
\end{equation}
followed by residual and feed-forward layers. For multi-view inputs, we apply the same attention to each view and aggregate the resulting tokens across views with element-wise max pooling, matching the main-text implementation. The output $\mathbf{H}\in\mathbb{R}^{B\times N_{\ell}\times C}$ is flattened as the fixed-size visual representation for the policy.

\subsection{Language-vision alignment.}
\label{app:lang-vis-align}

We convert $\mathbf{M}_{v}\in\mathbb{R}^{B\times C_v\times H\times W}$ into policy tokens with language-conditioned cross-attention, rather than simple token concatenation.
We flatten the map into $HW$ visual tokens, $\mathbf{X}_{v}\in\mathbb{R}^{B\times HW\times C_v}$, and add 2D RoPE positional encodings to preserve spatial coordinates~\cite{su2024roformer,heo2024rotary}.
In parallel, a frozen CLIP text encoder provides language tokens $\mathbf{X}_{\ell}\in\mathbb{R}^{B\times N_{\ell}\times C_{\ell}}$.
Lightweight adapters project both modalities to a shared dimension $C$:
$\hat{\mathbf{X}}_{v}=\psi_v(\operatorname{LN}(\mathbf{X}_{v}))\in\mathbb{R}^{B\times HW\times C}$ and
$\hat{\mathbf{X}}_{\ell}=\psi_\ell(\operatorname{LN}(\mathbf{X}_{\ell}))\in\mathbb{R}^{B\times N_{\ell}\times C}$.
We then perform multi-head cross-attention with language tokens as queries and visual tokens as keys/values:
\begin{equation}
\begin{aligned}
\tilde{\mathbf{H}}^{(m)}
&=\operatorname{MHA}\!\Big(
\text{query}=\hat{\mathbf{X}}_{\ell},\,
\text{key}=\hat{\mathbf{X}}_{v}^{(m)},\,
\text{value}=\hat{\mathbf{X}}_{v}^{(m)}
\Big) \\
&\in\mathbb{R}^{B\times N_{\ell}\times C},
\end{aligned}
\end{equation}
where $m$ indexes the input image/view (for a single image, set $m{=}1$).
Language queries focus computation on the instruction-relevant subspace while dense visual tokens remain keys and values.
For multi-image inputs, we compute $\tilde{\mathbf{H}}^{(m)}$ independently for each view and aggregate across views with element-wise max pooling:
\begin{equation}
\tilde{\mathbf{H}}=\max_{m\in\{1,\dots,M\}} \tilde{\mathbf{H}}^{(m)}\in\mathbb{R}^{B\times N_{\ell}\times C}.
\end{equation}

A standard Transformer block maps $\tilde{\mathbf{H}}$ to $N_{\ell}$ cross-attended tokens $\mathbf{H}\in\mathbb{R}^{B\times N_{\ell}\times C}$.
We flatten $\mathbf{H}$ and apply a small MLP to obtain the observation embedding consumed by the policy.

\section{Training, Benchmarks, and Deployment Protocols}
\label{app:experimental_details}

This section gives the shared training settings, benchmark protocols, task suites, runtime measurements, and physical deployment details used by the main experiments.

\begin{figure*}[t]
    \centering
    \includegraphics[width=\textwidth]{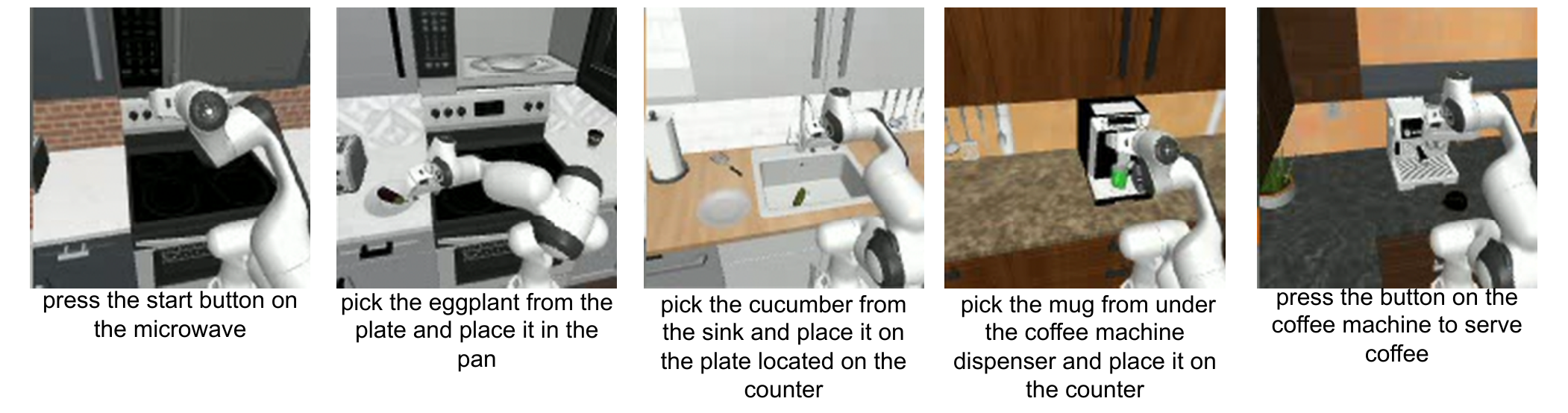}
    \caption{\textbf{RoboCasa tasks span semantic selection and contact-sensitive manipulation.}
    Representative examples from the 24-task suite are shown with their natural-language instructions.}
    \label{fig:Robocasa_tasks}
\end{figure*}

\begin{table}[t]
\centering
\footnotesize
\setlength{\tabcolsep}{5pt}
\renewcommand{\arraystretch}{1.10}
\caption{\textbf{Training hyperparameters for representation adaptation and policy learning.}
The table reports the settings used for the DROID adaptation stage and downstream policy optimization.}
\label{tab:training_hparams}
\begin{tabular}{@{}P{0.50\columnwidth} P{0.42\columnwidth}@{}}
\toprule
\textbf{Hyperparameter} & \textbf{Value} \\
\midrule
Effective global batch size & 256 \\
Optimizer & Adam \\
Learning rate & $10^{-4}$ \\
Learning-rate scheduler & Linear \\
\midrule
Observation processing MLP & $[1024,\,512,\,512]$ \\
Image resolution & $256{\times}256$ \\
\midrule
Diffusion method & DDIM \\
EMA power & 0.75 \\
1D U-Net hidden sizes & $[256,\,512,\,1024]$ \\
Observation horizon & 2 \\
Prediction horizon & 16 \\
Action horizon & 8 \\
\bottomrule
\end{tabular}
\end{table}

\begin{table}[t]
\centering
\footnotesize
\setlength{\tabcolsep}{5pt}
\renewcommand{\arraystretch}{1.10}
\caption{\textbf{Benchmark protocols for simulation and transfer evaluation.}
The table specifies the dataset split, training data, and rollout protocol used for each benchmark.}
\label{tab:benchmark_protocol}
\begin{tabular}{@{}P{0.34\columnwidth} P{0.24\columnwidth} P{0.32\columnwidth}@{}}
\toprule
\textbf{Component} & \textbf{Setting} & \textbf{Value} \\
\midrule
LIBERO (X-IL representation controls) & Epochs & 100 \\
LIBERO (X-IL representation controls) & Demos per task & 50 \\
LIBERO (X-IL representation controls) & Evaluation rollouts & 20 per task (not 20\% data) \\
LIBERO-10 deployment comparison & Evaluation rollouts & 50 per task \\
\midrule
RoboCasa (24-task encoder swaps) & Epochs & 100 \\
RoboCasa (24 tasks) & Evaluation episodes & 50 per task \\
\bottomrule
\end{tabular}
\end{table}

\subsection{DROID Adaptation and Downstream Policy Learning}
\label{app:implementation}

This section details the two-stage training protocol, the architectural initialization of the Teacher--Student framework, and the compute settings for teacher-anchored representation adaptation and downstream policy learning.

\subsubsection{Two-Stage Training Protocol}
\label{app:two_stage}
Our training follows a decoupled two-stage design to adapt large-scale pretrained models to robotic control.

\paragraph{Stage I: Teacher-Anchored Adaptation.}
We first adapt the diffusion-based representation on the DROID dataset~\cite{khazatsky2024droid}, which comprises 76k real-world manipulation episodes, to align the Stable Diffusion 2.1 (SD2.1) feature hierarchy with robotic viewpoints and manipulation imagery.
This stage optimizes the robot imitation objective together with the annealed feature-alignment loss in Eq.~\ref{eq:app-align}, so the Student representation is both policy-facing and regularized against the frozen Teacher manifold.
We follow the training protocol summarized in Table~\ref{tab:training_hparams}.
The resulting Student model retains the original SD2.1 module structure (VAE and U-Net), enabling it to serve as a versatile feature backbone for varied downstream tasks.

\paragraph{Stage II: Downstream Policy Learning.}
For simulation benchmarks including LIBERO and RoboCasa, the adapted Student encoder remains fixed.
We train the S2-FPN/readout modules and a diffusion policy with a 1D U-Net that operates over action trajectories, conditioning the policy on visual observations and language via cross-attention.
In this stage, the Student encoder is not updated; the policy-facing readout and diffusion policy are trained for the downstream task.
Key diffusion-policy hyperparameters and temporal horizons are provided in Table~\ref{tab:training_hparams}.

\subsubsection{Architectural Initialization and Optimization}
\label{app:initialization}
\paragraph{Configuration.} We use the SD2.1 U-Net as a frozen Teacher $f_{\theta_0}$. The Student branch is initialized by copying the Teacher weights for shared U-Net layers ($\theta \leftarrow \theta_0$), preserving the original convolutional scales and hierarchy. Student-side S2-FPN modules, readout adapters, and projection heads are initialized separately.
\paragraph{Optimization.} During Stage I, we update only Student-side parameters, including the Student U-Net, S2-FPN modules, policy-facing adapters, and projection heads $\{g_\phi^{(k)}\}$. The Teacher remains the original frozen model, and gradients are stopped on the Teacher branch so its features act as fixed manifold targets. In Stage II, the adapted Student encoder remains fixed, while S2-FPN/readout modules and policy layers are trainable.

\subsubsection{Adaptation Protocol and Hardware}
\label{app:distillation_protocol}
\paragraph{Noise Injection.} Following the SD2.1 schedule ($T=1000$), we sample timesteps $\tau \sim \mathcal{U}\{1,\ldots,T-1\}$ to construct noised latents:
\begin{equation}
z_\tau = \sqrt{\bar{\alpha}_\tau} z_0 + \sqrt{1 - \bar{\alpha}_\tau} \epsilon, \quad \epsilon \sim \mathcal{N}(0, \mathbf{I}),
\label{eq:noise_injection}
\end{equation}
where $z_0$ is the clean latent and $\epsilon$ is standard Gaussian noise. For multi-view inputs, we process each view independently and aggregate the resulting per-view alignment losses.
\paragraph{Preprocessing and Compute.} Images are resized to $256 \times 256$ pixels without data augmentation, maintaining consistency with standard diffusion-policy pipelines. DROID adaptation is performed for 300,000 steps on $8 \times$ NVIDIA A100 GPUs. Low-level hyperparameters, including the optimizer family and learning rate schedules, follow the defaults established in the DROID repository~\cite{khazatsky2024droid}.

\subsection{Simulation and Physical Evaluation Protocols}
\label{app:evaluation_protocols}

This section specifies the simulation benchmark settings, runtime profiling procedure, and real-robot protocol used by the main experiments.

\subsubsection{Simulation Benchmarks and Evaluation Protocols}
\label{app:sim_setup}

\paragraph{LIBERO Benchmark~\cite{liu2023libero}.}
We use LIBERO in two settings.
We follow the protocol used by prior work~\cite{jia2025x,li2025unified}: each suite contains 10 tasks, policies are trained for 100 epochs, and success is evaluated over 50 rollouts per task.
For the fixed-encoder representation controls in Fig.~\ref{fig:representation_interface_landscape}b, we use the full-data setting with 50 demonstrations per task and evaluate each task with 20 rollouts.
These controls are used to compare readout choices under a fixed policy protocol rather than to claim a new LIBERO leaderboard result.
For the deployment comparison, we include action-only Diffusion Policy variants~\cite{chi2025diffusion}, OpenVLA~\cite{kim2024openvla}, UniPi~\cite{du2023learning}, $\pi_0$ and $\pi_0$-FAST~\cite{black2410pi0}, and UVA~\cite{li2025unified}.
Inference time is reported for a single action trajectory.
All methods except \emph{OpenVLA} infer a 16-step sequence with 8 executed steps; \emph{OpenVLA} predicts one action per forward pass and is executed 8 times to match the same control horizon~\cite{kim2024openvla,li2025unified}.

\paragraph{RoboCasa Benchmark~\cite{nasiriany2024robocasa}.}
Following RoboCasa baselines, training and evaluation are performed on disjoint scene instances to assess generalization across diverse scene layouts and objects. We use a 24-task subset excluding navigation-dependent tasks. Unless otherwise specified, formal encoder-swap controls are trained for 100 epochs and evaluated over 50 episodes per task.
All formal representation controls use an effective global batch size of 256.
Multi-GPU execution is used only to reduce wall-clock time, and when an encoder requires more memory we keep the same effective batch size using gradient accumulation.

\paragraph{RoboCasa Task Specification.}
The selected 24-task subset spans pick-and-place, door operation, drawer operation, knob/lever turning, button pressing, and insertion-based coffee routines.
These tasks require both semantic grounding and pose-sensitive spatial control over articulated objects.
Fig.~\ref{fig:Robocasa_tasks} visualizes representative task instances, and Appendix Table~\ref{tab:robocasa-benchmark-full} lists the task names used for category-level reporting.

\subsubsection{Q4: Real-Robot Deployment Protocol}
\label{app:real_robot_details}

\begin{figure}[t]
    \centering
    \includegraphics[width=\columnwidth]{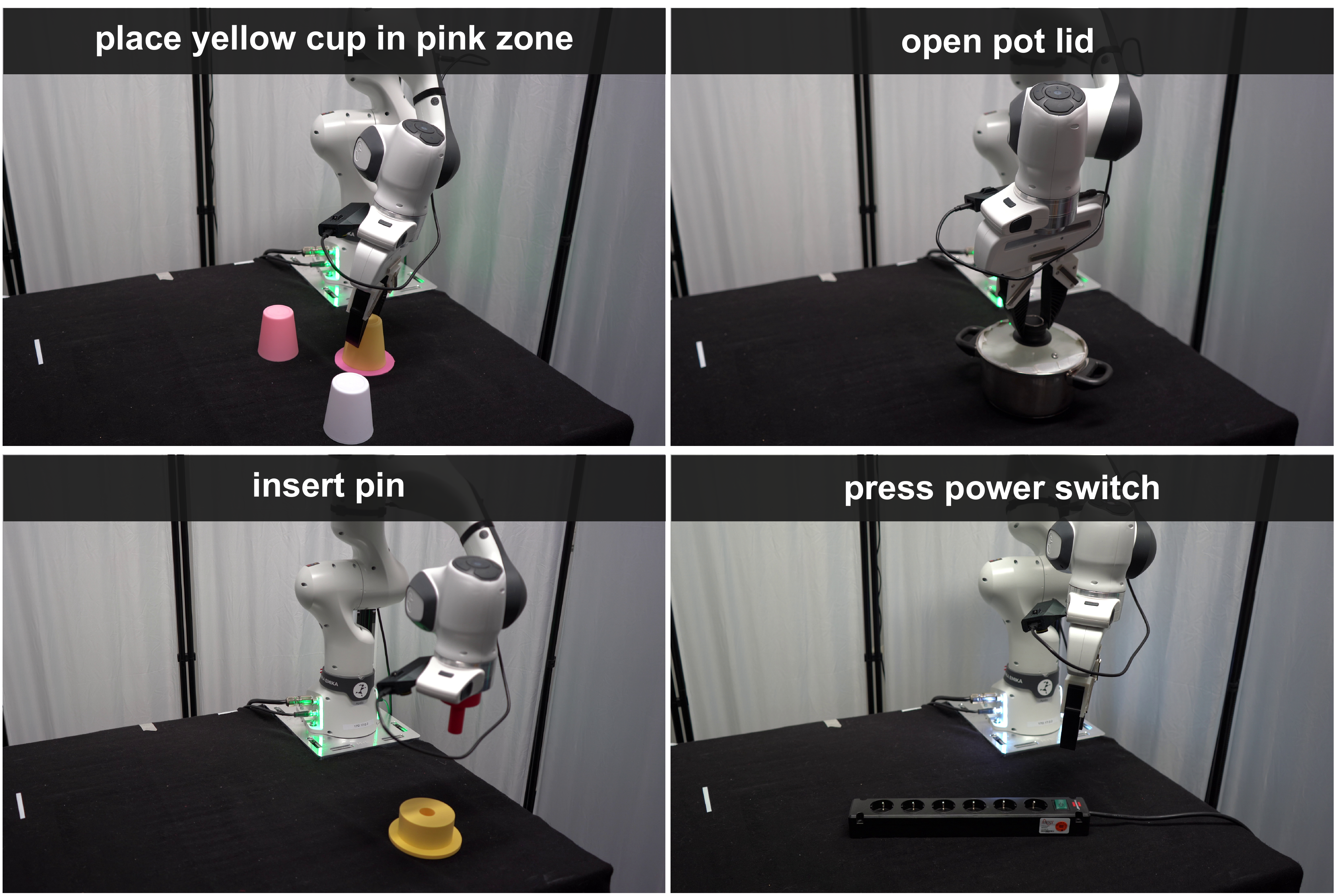}
    \caption{\textbf{Physical tasks probe increasing spatial precision.}
    The real-robot suite spans coarse semantic grounding (\emph{Sort Cup}, \emph{Open Lid}) and contact-rich alignment with tight tolerances (\emph{Insert Pin}, \emph{Press Switch}).}
    \label{fig:real_robot_task}
\end{figure}

All real-robot experiments are conducted on a 7-DoF Franka Emika Panda robot equipped with two ZED cameras: a wrist-mounted camera for hand--eye coordination and a third-person camera observing the workspace.
Object poses and target placements are reset at the beginning of each trial.
The four tasks are:
\emph{Sort Cup}, placing a yellow cup into a pink target zone among multiple cups;
\emph{Open Lid}, opening a pot lid under sustained contact;
\emph{Insert Pin}, a high-precision insertion task with approximately $2\,\mathrm{mm}$ clearance; and
\emph{Press Switch}, toggling a power switch with a small effective contact region ($<4\,\mathrm{cm}^2$).

All methods use the same Diffusion Policy implementation from \emph{DROID}~\cite{khazatsky2024droid}.
The visual encoder remains fixed for every method: internet-scale pretraining for DINOv2 and SigLIP, and DROID adaptation for Robot-DIFT.
The policy-facing readout and policy are trained under the same protocol.
We train the policy head for $10$K gradient steps on identical task datasets containing limited demonstrations per task: \emph{Sort Cup} (36), \emph{Open Lid} (31), \emph{Insert Pin} (35), and \emph{Press Switch} (35).
During deployment, each policy predicts a 16-step action sequence and executes the first 8 steps before replanning~\cite{zhao2023learning}.

\section{Appendix Evidence for the Main Experiment Claims}
\label{app:empirical_analysis}

This section provides the evidence behind the four experiment questions in Sec.~\ref{sec:experiments}.
It starts from the full RoboCasa encoder-swap table, then isolates readout effects, and finally reports Manifold Distillation and freeze-transfer diagnostics.
Unless stated otherwise, RoboCasa values are success rates over 50 rollouts per task.
Aggregate RoboCasa rows use completed 24-task evaluations only, and partial evaluations are excluded from main average comparisons.
The plotted values are drawn from the monitored experiment tables used for the main RoboCasa and LIBERO summaries.

\subsection{Q1: Controlled Encoder Swaps on RoboCasa}
\label{app:q1_encoder_swaps}

\paragraph{Full RoboCasa representation table.}
Table~\ref{tab:robocasa-benchmark-full} reports the per-task values summarized by category in Table~\ref{tab:robocasa-benchmark}.

\begin{table*}[p]
\centering
\caption{\textbf{Per-task RoboCasa results underlying the category-level benchmark.}
Success rates are measured over 50 rollouts per task under the same fixed-encoder protocol.
Bold marks the best score per task, including ties.
The contact-sensitive average combines \emph{Pressing Buttons} and \emph{Insertion}.}
\label{tab:robocasa-benchmark-full}
\footnotesize
\setlength{\tabcolsep}{1.8pt}
\renewcommand{\arraystretch}{1.10}
\resizebox{\textwidth}{!}{
\begin{tabular}{@{}llcccccccccc@{}}
\toprule
\multirow{2}{*}{\textbf{Category}} & \multirow{2}{*}{\textbf{Task}} &
\multicolumn{2}{c}{\textbf{Vision--language}} &
\multicolumn{2}{c}{\textbf{Geometry-oriented}} &
\multicolumn{2}{c}{\textbf{Self-supervised}} &
\multicolumn{2}{c}{\textbf{Generative}} &
\multicolumn{2}{c}{\textbf{DROID-ft}} \\
\cmidrule(lr){3-4}\cmidrule(lr){5-6}\cmidrule(lr){7-8}\cmidrule(lr){9-10}\cmidrule(lr){11-12}
& & \textbf{CLIP} & \textbf{SigLIP} & \textbf{VGGT} & \textbf{FastVGGT} & \textbf{DINOv2} & \textbf{DINOv3} & \textbf{SiT} & \textbf{DIFT} & \makecell[c]{\textbf{Robot}\\\textbf{DINO}} & \makecell[c]{\textbf{Robot}\\\textbf{DIFT}} \\
\midrule
\multirow{8}{*}{Pick-and-place}
& PnPCabToCounter       & \textbf{0.10} & 0.02 & \textbf{0.10} & \textbf{0.10} & \textbf{0.10} & 0.02 & 0.08 & 0.06 & 0.06 & \textbf{0.10} \\
& PnPCounterToCab       & 0.00 & 0.00 & 0.06 & \textbf{0.08} & 0.00 & 0.00 & 0.06 & 0.04 & \textbf{0.08} & \textbf{0.08} \\
& PnPCounterToMicrowave & 0.02 & 0.02 & 0.00 & 0.02 & 0.02 & 0.02 & 0.02 & 0.06 & 0.02 & \textbf{0.10} \\
& PnPCounterToSink      & 0.00 & 0.00 & 0.00 & 0.00 & 0.00 & 0.00 & \textbf{0.08} & \textbf{0.08} & 0.04 & 0.06 \\
& PnPCounterToStove     & 0.06 & 0.06 & 0.04 & 0.04 & 0.06 & 0.06 & 0.02 & 0.04 & \textbf{0.08} & \textbf{0.08} \\
& PnPMicrowaveToCounter & 0.00 & 0.02 & 0.06 & \textbf{0.08} & 0.00 & 0.04 & \textbf{0.08} & \textbf{0.08} & 0.06 & \textbf{0.08} \\
& PnPSinkToCounter      & 0.04 & 0.06 & 0.00 & 0.02 & 0.04 & 0.08 & 0.10 & 0.10 & \textbf{0.26} & \textbf{0.26} \\
& PnPStoveToCounter     & 0.00 & 0.00 & 0.04 & 0.04 & 0.00 & 0.02 & 0.08 & 0.04 & \textbf{0.20} & \textbf{0.20} \\
\midrule
\multirow{4}{*}{Doors}
& OpenSingleDoor        & 0.14 & 0.30 & 0.40 & 0.40 & 0.34 & 0.44 & 0.48 & 0.56 & \textbf{0.60} & 0.58 \\
& OpenDoubleDoor        & 0.00 & 0.10 & 0.10 & 0.10 & 0.04 & 0.04 & 0.38 & 0.18 & 0.30 & \textbf{0.92} \\
& CloseDoubleDoor       & 0.08 & 0.58 & 0.50 & 0.56 & \textbf{0.62} & 0.54 & 0.42 & 0.60 & 0.40 & 0.60 \\
& CloseSingleDoor       & 0.70 & 0.70 & 0.74 & 0.78 & 0.80 & 0.78 & 0.68 & 0.62 & 0.74 & \textbf{0.84} \\
\midrule
\multirow{2}{*}{Drawers}
& OpenDrawer            & 0.20 & 0.44 & 0.30 & 0.30 & 0.38 & 0.56 & 0.46 & \textbf{0.70} & 0.68 & 0.60 \\
& CloseDrawer           & 0.82 & 0.78 & 0.94 & 0.94 & 0.94 & \textbf{1.00} & 0.62 & \textbf{1.00} & 0.96 & \textbf{1.00} \\
\midrule
\multirow{2}{*}{Knobs}
& TurnOnStove           & 0.24 & 0.26 & 0.22 & 0.20 & 0.30 & 0.42 & 0.28 & \textbf{0.58} & 0.52 & 0.44 \\
& TurnOffStove          & 0.12 & 0.08 & 0.10 & 0.08 & 0.12 & \textbf{0.24} & 0.02 & 0.22 & 0.16 & 0.22 \\
\midrule
\multirow{3}{*}{Levers}
& TurnOnSinkFaucet      & 0.24 & 0.40 & 0.60 & \textbf{0.62} & 0.42 & 0.36 & 0.20 & 0.56 & 0.60 & \textbf{0.62} \\
& TurnOffSinkFaucet     & 0.50 & 0.62 & 0.20 & 0.20 & 0.60 & \textbf{0.78} & 0.50 & \textbf{0.78} & 0.74 & 0.74 \\
& TurnSinkSpout         & 0.32 & 0.44 & \textbf{0.50} & \textbf{0.50} & 0.32 & 0.40 & 0.34 & 0.40 & 0.34 & 0.48 \\
\midrule
\multirow{3}{*}{Buttons}
& CoffeePressButton     & 0.22 & 0.46 & 0.02 & 0.04 & 0.44 & 0.68 & 0.60 & 0.70 & 0.76 & \textbf{1.00} \\
& TurnOnMicrowave       & 0.22 & 0.44 & 0.12 & 0.08 & 0.46 & 0.64 & 0.52 & 0.66 & 0.66 & \textbf{0.82} \\
& TurnOffMicrowave      & 0.42 & 0.52 & 0.30 & 0.36 & 0.54 & 0.52 & 0.62 & 0.72 & 0.86 & \textbf{0.90} \\
\midrule
\multirow{2}{*}{Insertion}
& CoffeeServeMug        & 0.10 & 0.08 & 0.00 & 0.00 & 0.14 & 0.32 & 0.28 & 0.30 & 0.32 & \textbf{0.74} \\
& CoffeeSetupMug        & 0.02 & 0.04 & 0.00 & 0.00 & 0.02 & 0.02 & 0.02 & 0.10 & 0.08 & \textbf{0.18} \\
\midrule
\multicolumn{2}{c}{\textbf{Average}} &
0.19 & 0.27 & 0.22 & 0.23 & 0.28 & 0.33 & 0.29 & 0.38 & 0.40 & \textbf{0.49} \\
\multicolumn{2}{c}{\textbf{Contact-sensitive average}} &
0.20 & 0.31 & 0.09 & 0.10 & 0.32 & 0.44 & 0.41 & 0.50 & 0.54 & \textbf{0.73} \\
\bottomrule
\end{tabular}
}
\end{table*}

\begin{figure*}[t]
\centering
\includegraphics[width=\textwidth]{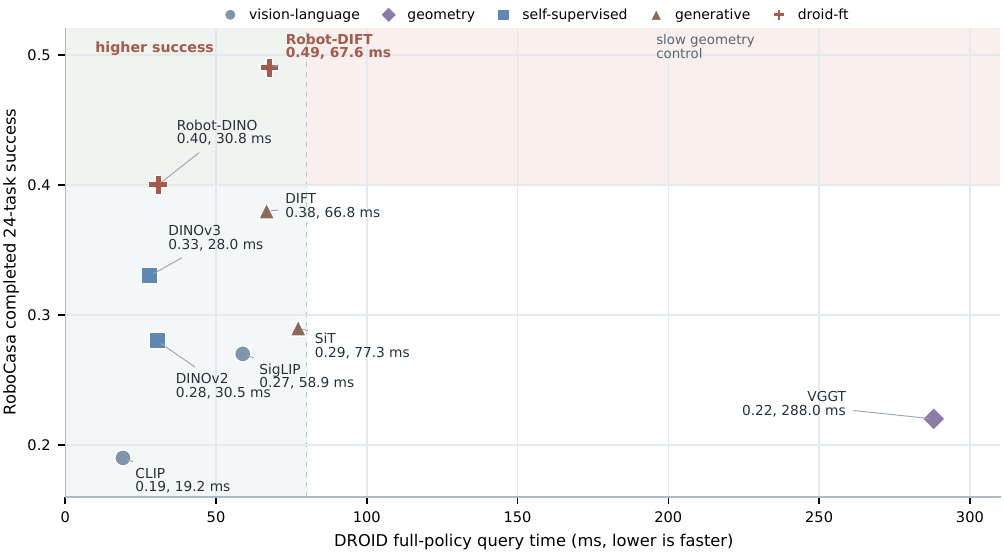}
\caption{\textbf{Accuracy--latency trade-off for fixed-encoder representation rows.}
Each point is a RoboCasa 24-task representation row evaluated under the same policy protocol.
Latency includes the fixed encoder and DroidDiffusionPolicy decoder under the predict-16 / execute-8 deployment chunk.
The vertical axis uses the task-weighted success reported in Table~\ref{tab:robocasa-benchmark}.}
\label{fig:corl_representation_evidence}
\end{figure*}

\paragraph{Full-policy time--success map.}
Appendix Fig.~\ref{fig:corl_representation_evidence} relates the completed 24-task RoboCasa rows from Table~\ref{tab:robocasa-benchmark} to DROID full-policy query latency.
This view is intended as a deployment-cost diagnostic rather than a separate benchmark.

\paragraph{DROID adaptation controls.}
Table~\ref{tab:droid_adaptation_controls} summarizes the completed RoboCasa full-table controls that separate robot-domain adaptation from representation choice.
Robot-DINO uses the same fixed-encoder transfer protocol as Robot-DIFT, but starts from a DINOv2+S2-FPN representation adapted on DROID rather than from a diffusion-derived hierarchy.
This control improves the DINO average only modestly, from 0.38 to 0.40, and shifts the task profile rather than uniformly improving contact-sensitive behavior.
By contrast, DROID adaptation of the diffusion-derived representation increases DIFT from 0.38 to 0.49 overall and from 0.50 to 0.73 on the contact-sensitive subset.
We interpret this as an interaction between the adaptation protocol and the pretrained feature structure: DROID data are useful when the interface can preserve dense local structure while aligning it to robot viewpoints, but robot-domain adaptation alone is not sufficient to make every pretrained representation equally actionable.

\begin{table*}[t]
\centering
\caption{\textbf{Robot-domain adaptation depends on the source representation.}
Rows report category-level RoboCasa success under the same fixed-encoder downstream policy protocol.
Robot-DINO matches the DROID adaptation setting but starts from DINOv2+S2-FPN.
Robot-DIFT starts from the diffusion-derived hierarchy and gains most on contact-sensitive categories.}
\label{tab:droid_adaptation_controls}
\footnotesize
\setlength{\tabcolsep}{3.0pt}
\renewcommand{\arraystretch}{1.08}
\resizebox{\textwidth}{!}{
\begin{tabular}{@{}lcccccccccc@{}}
\toprule
\textbf{Representation} & \textbf{DROID stage} & \textbf{Pick} & \textbf{Doors} & \textbf{Drawers} & \textbf{Knobs} & \textbf{Levers} & \textbf{Buttons} & \textbf{Insert.} & \textbf{Contact} & \textbf{Avg.} \\
\midrule
DINOv2+S2-FPN & No & 0.03 & 0.58 & 0.76 & 0.27 & 0.48 & 0.84 & 0.24 & 0.60 & 0.38 \\
Robot-DINO & Yes & 0.10 & 0.51 & 0.82 & 0.34 & 0.56 & 0.76 & 0.20 & 0.54 & 0.40 \\
\midrule
DIFT & No & 0.06 & 0.49 & 0.85 & 0.40 & 0.58 & 0.69 & 0.20 & 0.50 & 0.38 \\
\textbf{Robot-DIFT} & \textbf{Yes} & \textbf{0.12} & \textbf{0.74} & 0.80 & 0.33 & \textbf{0.61} & \textbf{0.91} & \textbf{0.46} & \textbf{0.73} & \textbf{0.49} \\
\bottomrule
\end{tabular}
}
\end{table*}

\subsection{Q2: Readout and Interface Controls}
\label{app:q2_readout_controls}

\begin{figure*}[t]
\centering
\includegraphics[width=\textwidth]{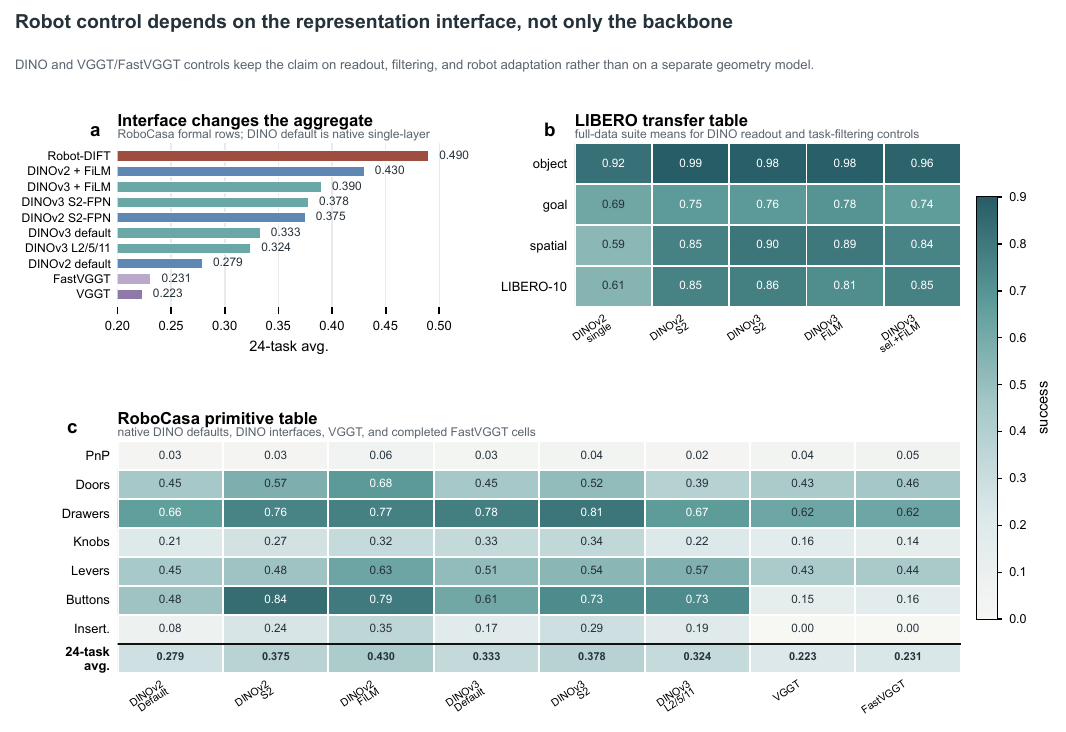}
\caption{\textbf{Policy-facing readout choices change how pretrained features transfer to control.}
Panel a compares RoboCasa 24-task averages for Robot-DIFT, native DINO defaults, DINO interface variants, and VGGT.
Panel b isolates the full-data LIBERO readout effect, including DINOv3 S2-FPN and selected-layer controls.
Panel c reports RoboCasa primitive success for DINO, VGGT, and FastVGGT readouts.
The pattern supports the view that spatial readout and task-conditioned filtering determine whether frozen features become actionable.}
\label{fig:representation_interface_landscape}
\end{figure*}

\paragraph{Representation-interface controls.}
Figure~\ref{fig:representation_interface_landscape} summarizes the aggregate interface ladder, full-data LIBERO controls, and primitive-level RoboCasa success table used to interpret the main encoder-swap comparison.
The figure is intended to make the task-dependent pattern visible rather than to introduce a new metric: no pretrained backbone or auxiliary cue is uniformly dominant.
DINO default rows use the native single-layer readout, and S2-FPN rows are reported separately as policy-facing interface variants.
The main readout claim is based on single-layer, direct-concat, S2-FPN, and geometry-control comparisons.
DINOv3 responds differently to S2-FPN fusion and selected-layer access.
VGGT is used here as a geometry-oriented control: it reaches 0.223 task-weighted 24-task average and remains much weaker than Robot-DIFT on the contact-sensitive subset.
FastVGGT reaches 0.231 task-weighted 24-task average and 0.096 on the contact-sensitive subset.
We use these rows to test whether VGGT-style spatial priors are directly actionable under the same fixed-encoder policy interface, without making the paper depend on a separate geometry model.
On LIBERO, replacing a single DINOv2 feature with S2-FPN improves all four full-data suites, with the largest gains on \emph{LIBERO-spatial} and \emph{LIBERO-10}.
The same monitored result tables are used for the main RoboCasa and LIBERO summaries.
Table~\ref{tab:robotdift_ablations}b in the main text adds a focused \emph{LIBERO-10} readout control using original, non-DROID-adapted DINOv2 and DIFT backbones.
Both backbones improve substantially when the readout moves from one layer to multi-layer access, and S2-FPN gives the strongest success for each backbone.

\paragraph{Latent-space diagnostics.}
We use six held-out LIBERO frames as a lightweight mechanism check for whether S2-FPN simply forwards one backbone layer or constructs a new policy-facing latent space.
These diagnostics are not used as performance evidence, but they support the interpretation of S2-FPN as a representation interface.
For DINOv3, the fused map has lower spatial entropy and higher peak-to-mean energy than raw layers, with only 0.354--0.426 linear centred kernel alignment (CKA) to the tested raw DINOv3 layers.
For the Robot-DIFT Student, the fused map also has the lowest tested entropy and 0.368--0.680 CKA to raw diffusion U-Net layers, closest to local decoder stages but not identical to any single one.
Cross-encoder diagnostics further show that DINOv3 and the Robot-DIFT Student share local spatial structure without collapsing to the same representation: DINOv3 layer 2 and Robot-DIFT us10 have CKA 0.788 and energy cosine 0.728, whereas the two S2-FPN fused maps have only moderate similarity, CKA 0.464 and energy cosine 0.450.
Caption perturbation further concentrates in the fused Robot-DIFT map: replacing the correct text with empty or wrong text produces relative feature changes of 1.439 and 1.433, with fused-map CKA dropping to 0.652 and 0.711.

\begin{figure*}[t]
\centering
\includegraphics[width=\textwidth]{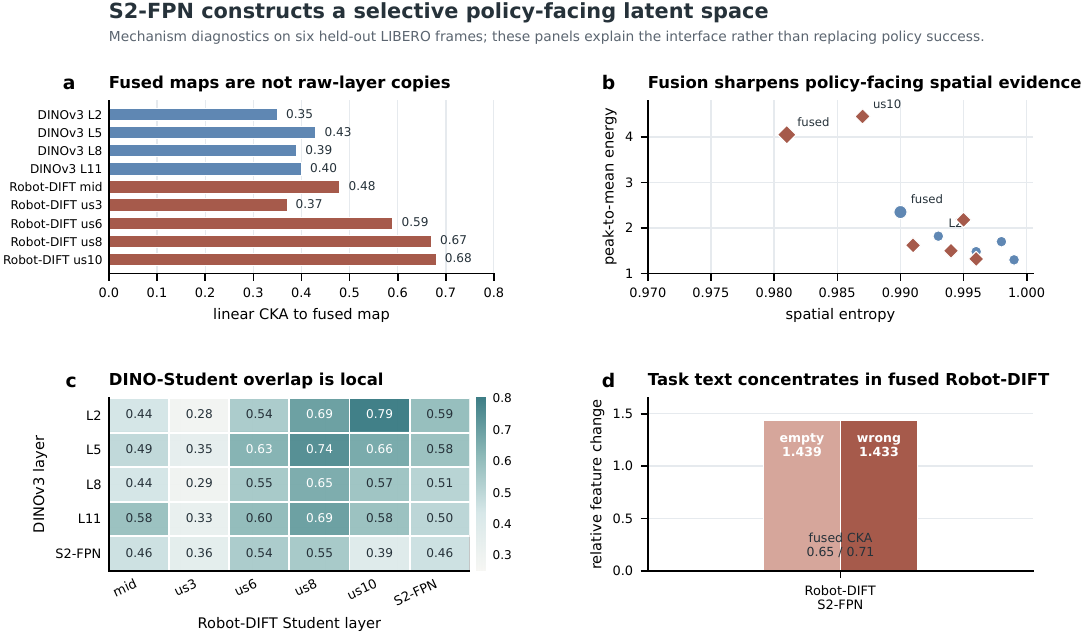}
\caption{\textbf{S2-FPN builds a policy-facing latent space rather than copying one raw layer.}
Diagnostics are computed on six held-out LIBERO frames and are used as mechanism support.
Panel a compares fused maps with raw DINOv3 and Robot-DIFT Student layers.
Panel b shows that fusion sharpens spatial activations.
Panel c compares cross-encoder structure, and panel d measures sensitivity to task text.}
\label{fig:latent_diagnostics}
\end{figure*}

\begin{figure*}[t]
    \centering
    \includegraphics[width=\textwidth]{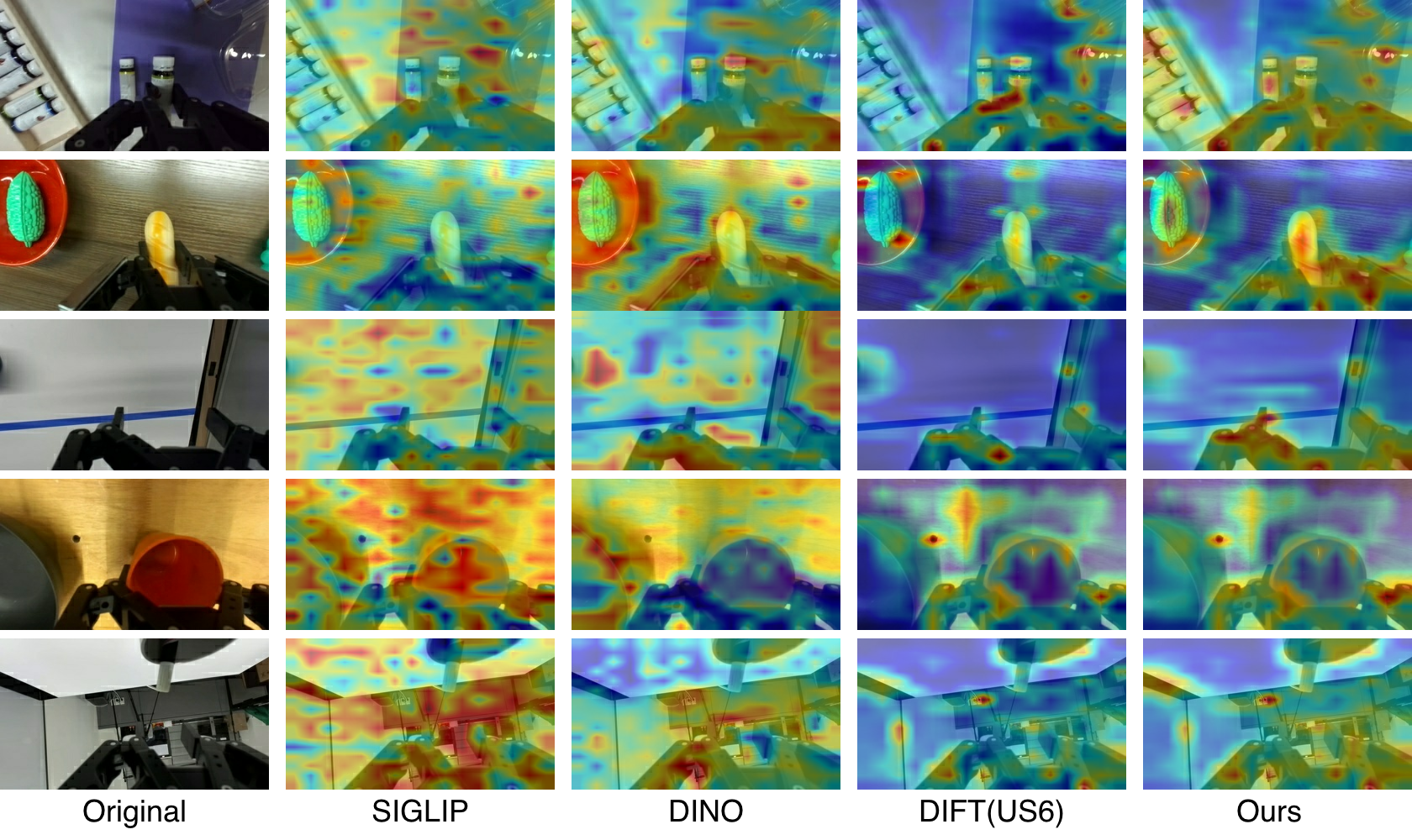}
    \caption{\textbf{Robot-DIFT localizes instruction-relevant robot-object regions.}
    Columns show the input image followed by SigLIP, DINO, DIFT, and Robot-DIFT attention or relevance maps.
    Rows show representative DROID real-world observations.
    Robot-DIFT produces more interaction-centric activations around targets and the end effector, with fewer background-dominated responses.}
    \label{fig:lang_attention}
\end{figure*}

\paragraph{Qualitative interaction evidence.}
Figure~\ref{fig:lang_attention} provides a qualitative check that the policy-facing features attend to robot-object interaction regions rather than only to object identity or background context.

\subsection{Q3: Manifold Distillation and Freeze-Transfer Diagnostics}
\label{app:q3_manifold_diagnostics}

\paragraph{Robot-DIFT component ablations.}
Table~\ref{tab:robotdift_ablations}a in the main text summarizes two implementation-level ablations on contact-sensitive RoboCasa tasks.
For multi-scale feature extraction, we first complete DROID adaptation, freeze the Student backbone, and train the same downstream policy for every variant.
\textit{SingleScale-us3}, \textit{SingleScale-us6}, and \textit{SingleScale-us8} each use one Student decoder feature map.
\textit{MultiScale (Ours)} instead fuses $\{\mathbf{s}^{(us3)}, \mathbf{s}^{(us6)}, \mathbf{s}^{(us8)}\}$ with S2-FPN.
The single-scale variants expose a trade-off between global context and local spatial detail.
Coarser features retain task-level context but under-resolve local contact regions, while finer features preserve local structure but are more sensitive to distractors.
The multi-scale readout gives the policy both signals and achieves the highest mean success in the contact-sensitive subset.

For alignment scheduling, \textit{NoAnneal} keeps the alignment weight fixed throughout training, while \textit{Anneal (Ours)} follows the decayed alignment schedule in Sec.~\ref{sec:manifold_distillation}.
The comparison tests whether a time-varying constraint better balances preservation of the pretrained diffusion feature manifold and robot-domain specialization.
Annealing improves both pressing-button and insertion tasks, indicating that strong early alignment stabilizes the Student while later relaxation allows task-specific adaptation.

\begin{figure}[t]
    \centering
    \includegraphics[width=\columnwidth]{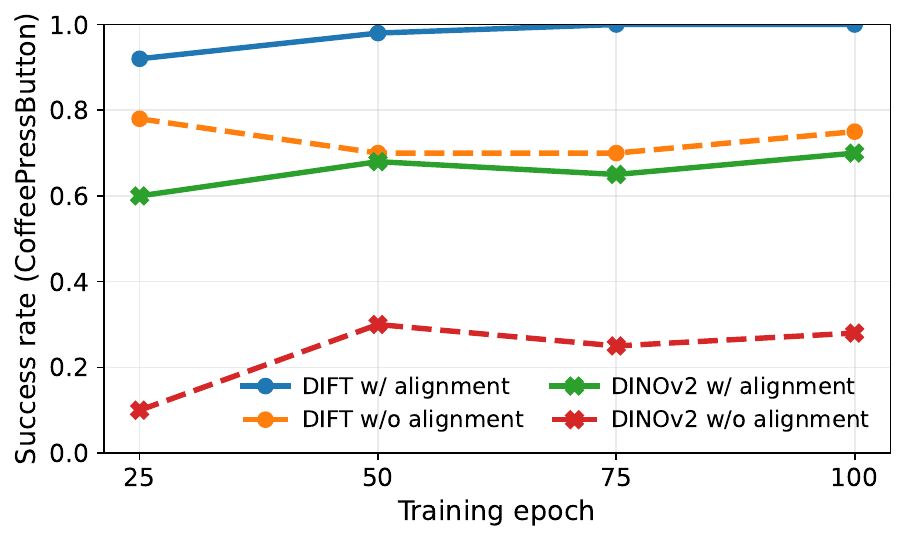}
    \caption{\textbf{Alignment stabilizes finetuning on a contact-sensitive RoboCasa task.}
Success is reported at epochs 25/50/75/100 for DIFT and DINOv2 with or without the alignment loss.
Including alignment improves final success and reduces late-training degradation in this CoffeePressButton setting.}
\label{fig:align_ablation_coffee}
\end{figure}

\paragraph{Impact of the Alignment Loss.}
To isolate the contribution of the alignment objective, which regularizes the adapted representation against the pretrained feature manifold, we perform an ablation on the RoboCasa \texttt{CoffeePressButton} task.
We compare finetuning with and without the alignment loss across two backbones: DIFT (diffusion-based) and DINOv2 (discriminative).
Fig.~\ref{fig:align_ablation_coffee} reports success rates at multiple training checkpoints.
In both settings, the alignment loss improves final success and reduces late-training degradation.
This pattern is consistent with the alignment loss limiting drift away from the pretrained feature structure during closed-loop policy training.

\begin{figure*}[t]
    \centering
    \includegraphics[width=0.325\textwidth]{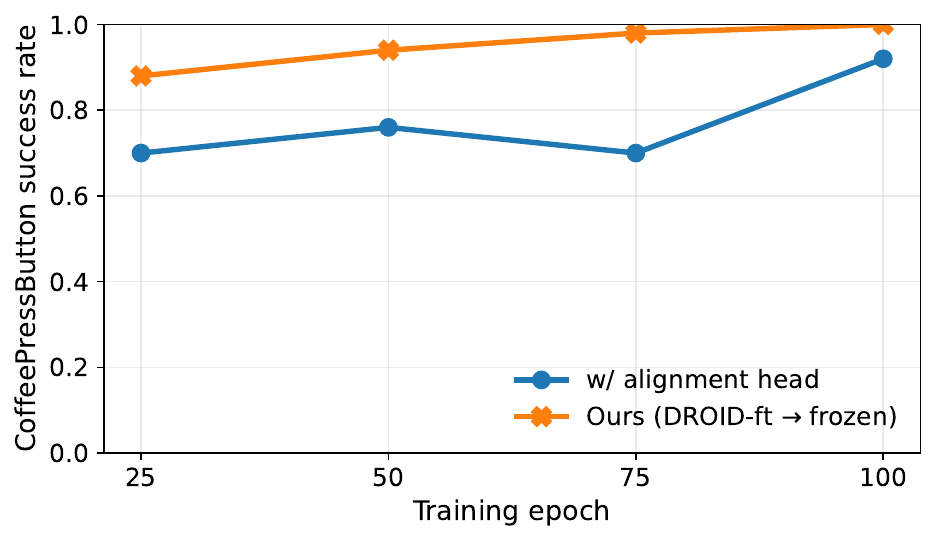}
    \includegraphics[width=0.325\textwidth]{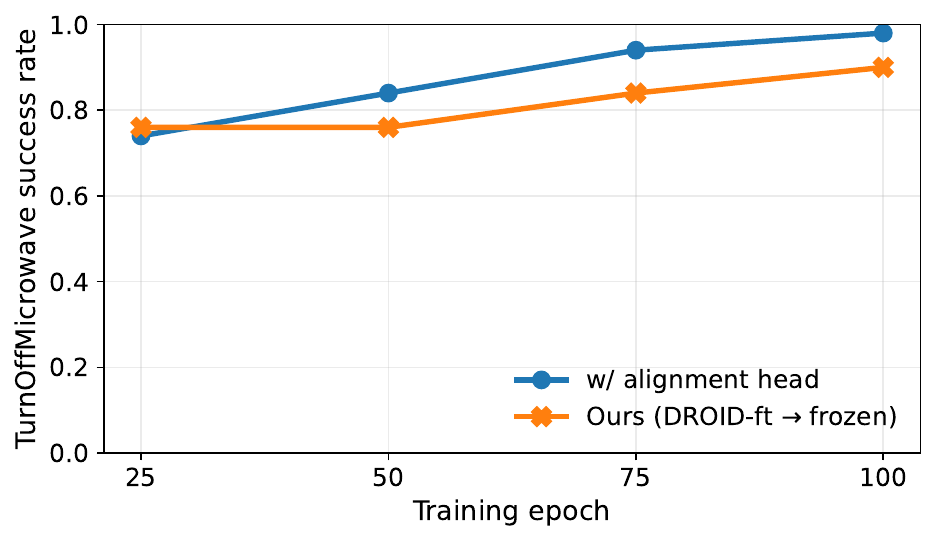}
    \includegraphics[width=0.325\textwidth]{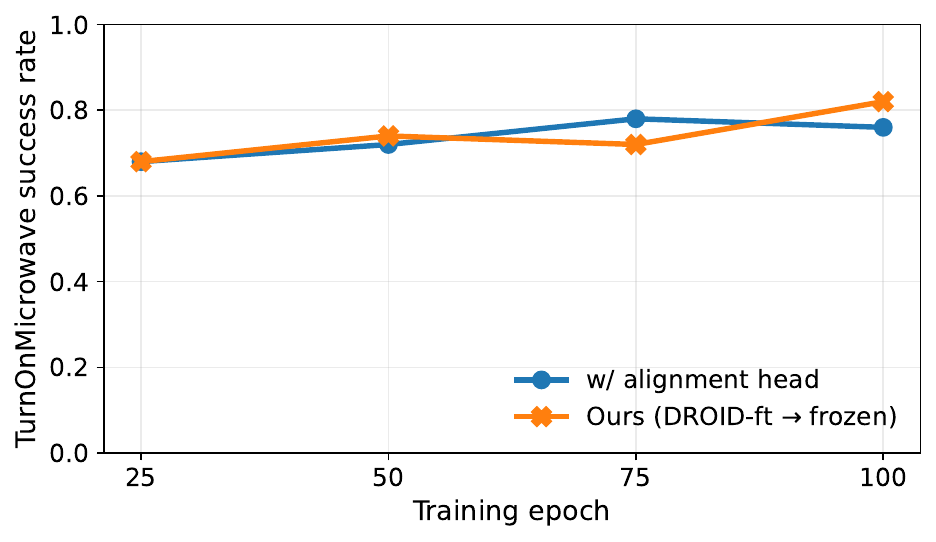}

\caption{\textbf{DROID adaptation transfers efficiently to RoboCasa with a fixed Student encoder.}
We compare a DROID-adapted representation transferred with the Student encoder fixed against direct RoboCasa finetuning with alignment.
Curves show success at epochs 25/50/75/100 for \texttt{CoffeePressButton}, \texttt{TurnOffMicrowave}, and \texttt{TurnOnMicrowave}.
The epoch-100 marker reports the final post-training evaluation.}
\label{fig:transfer_freeze_vs_finetune}

\end{figure*}

\paragraph{Transfer Learning from DROID to RoboCasa.}
We evaluate whether teacher-anchored adaptation on large-scale real-world data (DROID) provides a stronger initialization for simulation benchmarks than direct in-domain finetuning.
We compare the DROID-adapted fixed Student encoder with direct RoboCasa finetuning under the same alignment objective.
Fig.~\ref{fig:transfer_freeze_vs_finetune} reports success trajectories on three representative tasks: \texttt{CoffeePressButton}, \texttt{TurnOffMicrowave}, and \texttt{TurnOnMicrowave}.
The DROID-adapted encoder converges faster and reaches higher or comparable final success rates, indicating that teacher-anchored real-world adaptation yields transferable features and reduces the need for extensive in-domain finetuning.

\section{Appendix Takeaway}
\label{app:summary}

The appendix is organized to make the main-text claims auditable.
The method sections define how Robot-DIFT preserves a diffusion-derived feature hierarchy while exposing it through S2-FPN.
The protocol sections fix the evaluation settings used by RoboCasa, LIBERO, and the physical robot experiments.
The evidence sections then follow the main experiment logic: controlled encoder swaps establish the representation-choice result; readout controls show why the interface matters; and Manifold Distillation diagnostics test why the adapted Student must remain anchored to the pretrained feature distribution.

%% file: sections/2_related_works.tex
\section{Related Work}
\label{sec:related_works}

\subsection{Visual Representations for Visuomotor Policy Learning}
Early work in imitation learning (IL) and reinforcement learning (RL) typically trained shallow convolutional networks or ResNet-based encoders from scratch on task-specific data, achieving strong performance in constrained environments but limited robustness to visual distractors, or novel objects~\cite{chi2025diffusion, reuss2023goal, reuss2024multimodal, jia2024mail, mandlekar2021what, parisi2022unsurprising}.
To improve sample efficiency and generalization, subsequent approaches adopted pre-trained visual representations.
Self-supervised objectives such as masked reconstruction (e.g., MVP) and contrastive learning in egocentric or manipulation-centric datasets have demonstrated improved robustness over learning from scratch~\cite{xiao2022masked, nair2022r3m, lin2022egocentric, mandlekar2021what}.
However, these representations often require domain-specific pretraining or fine-tuning and do not naturally scale to the visual diversity captured by internet-scale data.
More importantly, their training objectives often favor invariance over the correspondence-sensitive local variation required for contact-rich manipulation.

\subsection{Foundation Visual Models and Vision--Language--Action Policies}
Recent progress has been driven by foundation visual models integrated into Vision--Language--Action (VLA) systems~\cite{black2410pi0, intelligence2504pi0, bjorck2025gr00t, reuss2025flower, li2024towards, kim2025fine}.
Vision--language backbones such as CLIP~\cite{radford2021learning} and SigLIP~\cite{zhai2023sigmoid}, and self-supervised visual backbones such as DINOv2~\cite{oquab2023dinov2}, are widely used as frozen encoders in generalist policies, enabling semantic grounding and instruction following~\cite{kim2024openvla, team2024octo}.
Similarly, RT-1 and RDT-1B rely on pretrained discriminative encoders (e.g., EfficientNet or SigLIP) to process visual observations~\cite{brohan2022rt, liu2024rdt}.
While highly effective at recognizing \emph{what} an object is, these representations often prioritize semantic invariance, which can underexpose the local correspondence cues needed for millimeter-level manipulation.
Empirical studies suggest that such representations may not expose dense spatial fidelity through standard policy interfaces for precise contact and pose reasoning~\cite{nair2022r3m}.
Attempts to compensate via explicit depth fusion or point-cloud backbones improve geometry but introduce additional computational overhead and sensor dependence~\cite{jia2025pointmappolicy, donat2025towards, haldar2025point}.
More recently, geometry-oriented foundation models such as VGGT~\cite{wang2025vggt} and FastVGGT~\cite{shen2026fastvggt} recover dense metric or multi-view 3D structure from RGB sequences, offering a complementary spatial inductive bias.
However, metric geometry reconstruction is not equivalent to the \emph{correspondence sensitivity} required for contact-rich control: policy-facing features must respond locally and repeatably to small pose and contact-layout shifts, rather than globally reconstruct scene geometry.
Robot-DIFT targets this finer-grained property by redirecting the correspondence-rich spatial hierarchy of diffusion features toward reactive closed-loop control.

\subsection{Generative Diffusion Models for Perception and Control}
Generative diffusion models provide a fundamentally different representational bias~\cite{rombach2022high,song2020score,dhariwal2021diffusion}.
Trained via iterative denoising, diffusion models are forced to preserve fine-grained spatial structure, and recent analyses reveal that their U-Net activations encode dense pixel-level correspondences and hierarchical part--whole relationships~\cite{zhang2023tale, stracke2025cleandift}.
DIFT shows that diffusion features contain strong semantic and dense spatial correspondence cues~\cite{zhang2023tale}.
CleanDIFT further demonstrates that these representations can be distilled into deterministic descriptors from clean inputs through layer-wise teacher--student alignment, outperforming discriminative features on correspondence tasks~\cite{stracke2025cleandift}.
In robotics, however, diffusion models have primarily been used to model the \emph{action} distribution, as in Diffusion Policy and RDT-1B, while perception continues to rely on standard discriminative encoders~\cite{chi2025diffusion, liu2024rdt}.
To our knowledge, no prior work has studied diffusion feature distillation as a \emph{robot-domain-distilled, single-pass, policy-facing} visual backbone for real-time manipulation.
Robot-DIFT builds on the correspondence-rich spatial structure of diffusion-based representations, but differs in both setting and objective: rather than using diffusion features as image-matching descriptors, we adapt diffusion feature distillation to closed-loop visuomotor policy learning, where inference must be single-pass and training must remain stable under policy-induced representation drift.